\documentclass[10pt]{article} %
\usepackage[accepted]{tmlr}

\usepackage{amsmath,amsfonts,bm}

\def\eqref#1{equation~\ref{#1}}

\def\1{\bm{1}}

\DeclareMathAlphabet{\mathsfit}{\encodingdefault}{\sfdefault}{m}{sl}
\SetMathAlphabet{\mathsfit}{bold}{\encodingdefault}{\sfdefault}{bx}{n}

\usepackage{hyperref}
\hypersetup{colorlinks,%
citecolor=gray,%
filecolor=gray,%
linkcolor=gray,%
urlcolor=gray
}
\usepackage{url}

\usepackage{microtype}
\usepackage{graphicx}
\usepackage{booktabs} %
\usepackage{caption}
\usepackage{wrapfig}
\usepackage{multirow}
\usepackage{multicol}
\usepackage{url}

\usepackage{amsfonts}
\usepackage{bm}
\usepackage{amsmath}
\usepackage{amssymb}
\usepackage{mathtools}
\usepackage{amsthm}

\usepackage{makecell}
\usepackage{colortbl}
\usepackage{enumitem}

\definecolor{emerald}{rgb}{0.31, 0.78, 0.47}
 
\definecolor{caribbeangreen}{rgb}{0.0, 0.8, 0.6}
\definecolor{green(munsell)}{rgb}{0.0, 0.66, 0.47}

\newcommand{\revise}[1]{{\color{black}#1}}

\title{What Does Softmax Probability Tell Us about \\
Classifiers Ranking Across Diverse Test Conditions?}

\author{\name Weijie Tu \email weijie.tu@anu.edu.au \\
      Australian National University
      \AND
      \name Weijian Deng \email weijian.deng@anu.edu.au \\
      Australian National University
      \AND
      \name Liang Zheng \email liang.zheng@anu.edu.au\\
      Australian National University
      \AND
      \name Tom Gedeon \email tom.gedeon@curtin.edu.au\\
      Australian National University \\
      Curtin University \\
      University of \'Obuda
      }

\def\month{05}  %
\def\year{2024} %
\def\openreview{\url{https://openreview.net/forum?id=vtiDUgGjyx}} %

\begin{document}

\maketitle

\begin{abstract}

This work aims to develop a measure that can accurately rank the performance of various classifiers when they are tested on unlabeled data from out-of-distribution (OOD) distributions. We commence by demonstrating that conventional uncertainty metrics, notably the maximum Softmax prediction probability, possess inherent utility in forecasting model generalization across certain OOD contexts. Building on this insight, we introduce a new measure called Softmax Correlation (SoftmaxCorr). It calculates the cosine similarity between a class-class correlation matrix, \revise{constructed} from Softmax output vectors across an unlabeled test dataset, and a predefined reference matrix that embodies ideal class correlations. A high resemblance of predictions to the reference matrix signals that the model delivers confident and uniform predictions across all categories, reflecting minimal uncertainty and confusion. Through rigorous evaluation across a suite of datasets, including ImageNet, CIFAR-10, and WILDS, we affirm the predictive validity of SoftmaxCorr in accurately forecasting model performance within both in-distribution (ID) and OOD settings. Furthermore, we discuss the limitations of our proposed measure and suggest avenues for future research.

\end{abstract}

\section{Introduction}
\label{sec:intro}

Machine learning (ML) models typically excel on test sets coming from the same distribution as the training set. However, this assumption seldom holds in real-world deployments, where the test environments often experience distribution shifts caused by factors such as sample bias and non-stationarity.  Recognizing this challenge, there is a pressing need to assess ML model performance in unlabeled testing environments where traditional evaluation metrics \revise{which require data annotations, such as accuracy and $F1$ score, may become infeasible.} Hence, our focus shifts towards the vital yet under-explored task of ranking models under these conditions. Specifically, given a pool of trained models and an unlabeled test set, the objective is to efficiently select the most suitable model for utilization.

Various complexity measures have been proposed to predict the in-distribution (ID) accuracy of models
\citep{jiang2020neurips,neyshabur2015norm, bartlett2017spectrally, keskar2016large, nagarajan2019generalization,neyshabur2017exploring,chuang2021measuring,jiang2019fantastic,smith2017bayesian, arora2018stronger, dziugaite2017computing, dinh2017sharp}.
For the OOD test sets, classical domain adaptation theory provides a partial answer by quantifying the distance between the ID and OOD distributions \citep{ben2006analysis}. 
Moreover, the ``accuracy-on-the-line" phenomenon~\citet{miller2021accuracy} shows the linear correlation in the probit scale between ID and OOD performance. However, such phenomenon does not always hold on some distributions~\citep{teney2024id}.
Building on previous research, our objective is to develop a robust measure capable of ranking models on both ID and OOD datasets \textit{without} testing labels.

Softmax prediction probability has been shown to be useful in analyzing test data in several tasks, such as open-set data detection~\citep{hendrycks2016baseline}, accuracy estimation~\citep{guillory2021predicting,garg2022leveraging}, and misclassified input detection~\citep{Deng2022trust}.
For example, \citet{hendrycks2016baseline} and \citet{liang2018enhancing} utilize maximum Softmax prediction probability to identify samples from open-set classes. 
Driven by these insights, we develop OOD measures based on Softmax probability. Concretely, given various deep models, we aim to develop probability-based measures that monotonically relate to OOD generalization.
To validate the feasibility, we conduct extensive and large-scale correlation studies using various models and different types of dataset shifts.
We construct a catalog of empirical prediction probability-based measures and create a wide range of experimental setups.
We collect $573$ different classification models ranging from standard convolutional neural networks to Vision Transformers.
We cover $11$ ID and OOD datasets with various types of distribution shift, such as ImageNet-V2 \citep{recht2019imagenet} with dataset reproduction shift and ImageNet-R \citep{hendrycks2021many} with style shift.

Based on experimental results, we first show the demand for measures to rank model performance beyond in-distribution accuracy \citep{miller2021accuracy}. Then, we observe that empirical uncertainty measures based on prediction probabilities (\emph{e.g.}, maximum softmax probability) are useful in characterizing OOD generalization to some extent. However, we note that these measures solely account for prediction certainty. \revise{In response, we introduce SoftmaxCorr, a method designed to harness both prediction certainty and diversity. Confidence pertains to the certainty of individual predictions, whereas dispersity indicates the spread of predictions across all categories.}
Specifically, for each classifier, we compute a class-class correlation matrix from all prediction vectors in a test set. Then, we calculate its cosine similarity with a predefined reference matrix, designed to represent desirable prediction patterns, to evaluate whether this classifier makes diverse and certain predictions. The broad correlation study shows the efficacy of SoftmaxCorr in ranking models.

\section{Related Work}
\label{sec:related_work}

\textbf{Predicting generalization in deep learning} studies the ID generalization gap (\emph{i.e.}, the difference between training and test accuracy) of deep neural networks. Representative methods develop complexity measures based on model parameters and training set \citep{jiang2020neurips,jiang2019fantastic,neyshabur2015norm, keskar2016large, bartlett2017spectrally, neyshabur2017pac, liang2019fisher, chuang2021measuring, smith2017bayesian, arora2018stronger, dziugaite2017computing, dinh2017sharp, dziugaite2020search}, such as distance of training weights from initialization \citep{nagarajan2019generalization}, {the product of norms of weights across layers} \citep{neyshabur2017exploring} and {the change of model accuracy with respect to different perturbation levels in training data} \citep{schiff2021predicting}. 
The above methods assume that training and test data come from the same distribution and do not incorporate characteristics of test data, so we can unlikely make reasonable predictions on a different distribution. 
To mitigate this limitation, we investigate the model generalization under distribution shift by developing measures that reflect the models' generalization ability on OOD datasets.

\paragraph{OOD generalization.} 
Machine learning models should generalize from training distribution to OOD datasets~\citep{djolonga2021robustness,koh2021wilds, kirsch2022note}. To study this problem, several benchmarks are proposed~\citep{hendrycks2019benchmarking,koh2021wilds,gulrajani2020search}, such as corruption benchmark \citep{hendrycks2019benchmarking} and domain generalization testbed~\citep{gulrajani2020search}.
Moreover, several methods are proposed to improve model OOD generalization~\citep{volpi2018generalizing,zhao2020maximum,sagawa2019distributionally,pmlr-v139-liu21f,mansilla2021domain,shi2021gradient, krishnamachari2023mitigating, zhang2021can, huang2023winning, foret2020sharpness, huang2022harnessing}, such as adversarial domain augmentation~\citep{volpi2018generalizing,qiao2021uncertainty, alhamoud2022generalizability} and inter-domain gradient matching~\citep{shi2021gradient}.

\revise{Prior research has established theoretical frameworks for assessing classifier performance amid distribution shifts. \cite{ben2006analysis} introduced the first VC-dimension-based generalization bound, quantifying the difference in classifier error between source and target distributions using a divergence measure. \cite{mansour2009domain} extended this analysis to more general loss functions, providing refined generalization bounds via Rademacher complexity. Other studies~\citep{blitzer2007learning, hoffman2018algorithms, mansour2008domain} expanded upon these findings to include multiple source domains. Some works further bound the OOD generalization error based on the divergence between the two distributions~\citep{acuna2021f, zhang2019bridging, tachet2020domain}
However, as suggested by \citet{miller2021accuracy}, when the distribution shift becomes larger, the above bounds on OOD performance become looser.
In addition,~\citet{vedantam2021empirical} report that the adapting theory from domain adaptation is limited in predicting OOD generalization. 
}

\paragraph{Unsupervised accuracy estimation (UAE)} aims to predict the performance of a given model on various unlabelled out-of-distribution datasets. 
One line of research utilizes the model outputs on the test sets \citep{guillory2021predicting, garg2022leveraging, deng2023confidence}. For instance, \citet{guillory2021predicting} uses the maximum value of prediction probability to estimate model accuracy. 
A parallel line of works predicts model performance by gauging distribution discrepancy between training and test sets using image features \citep{deng2021labels, Tu_2023_CVPR, xie2024importance, lu2024characterizing}. For example, \citet{deng2021labels} uses the first- and second-order statistics of image features and Fr\'echet Distance to jointly measure the model performance.
In contrast, other studies investigate this task through model weights \citep{yu2022predicting}.

\revise{While UAE predicts OOD accuracy of a single model across various test sets, our work focuses on ranking multiple models' OOD performance on an unlabeled test dataset. UAE methods may be deployed in this task by predicting the performance of each single model. However, to derive the estimated performance, most UAE methods (\emph{e.g.}, BoP \citep{Tu_2023_CVPR} or Dispersion score \citep{xie2024importance}) require training an accuracy predictor on a number of datasets. These operations make the deployment of methods computational expensive in our tasks. We have included efficient Softmax-based UAE methods in our study 
}

\section{Task Formulation}
\label{sec:formulation}

\textbf{Task definition.} We consider a $K$-way classification task, and let $\mathcal{Y}=\{1,...,K\}$ denote the label space and $\mathcal{X} \in \mathbb{R}^d$ denote the input space. 
We are given a labeled training set $\mathcal{D}^S$ 
that contains data i.i.d drawn from a source distribution $P_S$, and an OOD test set $\mathcal{D}^T \coloneqq \{ (x_i, y_i)\}_{i=1}^{N}$ that contains $N$ data i.i.d drawn from another distribution $P_T$ ($P_S \neq P_T$).
We train $M$ neural network classifiers $\{\phi_m\}_{m=1}^{M}$ on $\mathcal{D}^S$. Given a sample $(\bm{x},y)$ from $\mathcal{D}^T$, the classifier $\phi_{m}:\mathcal{X} \rightarrow \Delta^K$ gives Softmax probabilities for $\bm{x}$ on $K$ classes, where $\Delta^K$ denote $K-1$ dimensional unit simplex. 
By testing on $\mathcal{D}^T$, $\phi_m$ yields a prediction matrix~$\bm{P} \in \mathbb{R}^{N \times K}$, whose each row represents prediction probabilities of a test data. Specifically, the prediction matrix satisfies $\sum_{j=1}^K{{P}_{i, j}} = 1\;\, \forall i \in 1 \dots N$ and ${P}_{i, j} \geq 0\;\, \forall i \in 1 \dots N, j \in 1 \dots K$, where ${P}_{i, j}$ indicates the probability that $i$-th sample is predicted to the $j$-th class.

The dataset has an evaluation metric (\emph{e.g.}, accuracy) to obtain ground-truth generalization $G_m$ of classifier $\phi_m$. The goal is to design a measure to calculate a score $S_m$ for each classifier $\phi_m$ without access to data annotations. The calculated scores $\{S_m\}_{m=1}^{M}$ ideally should correlate well with $\{G_m\}_{m=1}^{M}$, so that we can rank the OOD generalization of models based on the scores.

\paragraph{Evaluation metrics.} We use Spearman's Rank Correlation coefficient $\rho$ \citep{kendall1948rank} to measure monotonicity between calculated scores and model generalization. In addition, we also compute the weighted variant of Kendall’s rank correlation $\tau_w$, which is shown to be a useful measure when selecting the best-ranked item of interest \citep{you2021logme}. Both coefficients range from $[-1, 1]$. A value closer to $-1$ or $1$ indicates a strong negative or positive correlation, respectively, and $0$ means no correlation. Similar to \citet{miller2021accuracy} and \citet{baek2022agreement}, we apply the same probit scale to both accuracy and SoftmaxCorr in our experiment for a better linear fit.

\section{Softmax Probability-based OOD Measures} \label{sec:measures}

\subsection{What Makes OOD Measures Interesting?} 
\begin{figure}
  \begin{center}
    \includegraphics[width=\linewidth]{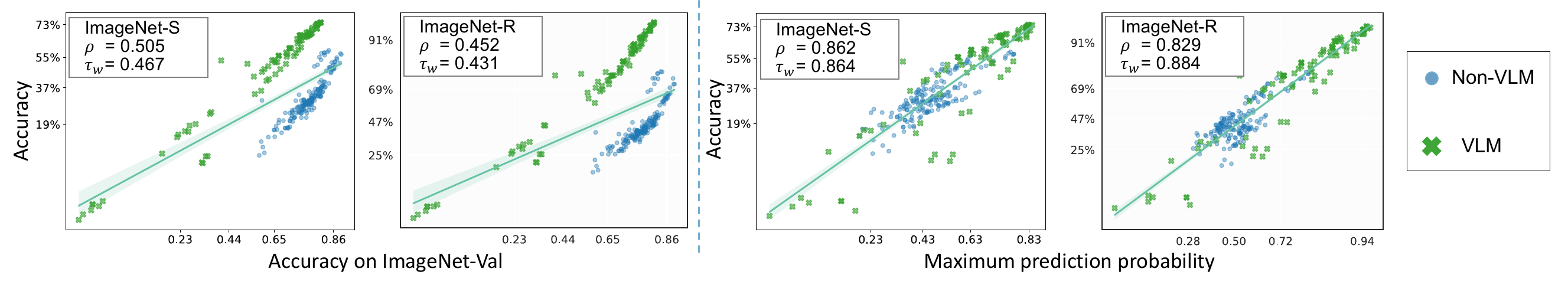}
      \vspace{-25pt}
    \caption{\textbf{Correlation study between MaxPred and accuracy (\%) on ImageNet-S and ImageNet-R}. Every point denotes a classifier. 
    \revise{We use $173$ ImageNet models and $89$ vision--language models introduced in Section~\ref{sec:experiment}. The straight line is fit with robust linear regression \citep{huber2011robust} and the shadow means the 95\% Clopper-Pearson confidence intervals. }
    We show that MaxPred exhibits a moderate correlation with accuracy, while accuracy on ImageNet-validation shows a relatively low correlation with performance on ImageNet-R.
    Moreover, Vision-Language Models (VLMs) exhibit varying linear trends in terms of their ID and OOD accuracy compared to standard supervised models.}
    \label{fig:maxpred}
  \end{center}
  \vspace{-15pt}
\end{figure}

\paragraph{Beyond Accuracy-on-the-Line (AoL).} \citet{miller2021accuracy} report an AoL phenomenon where there exists a strong linear correlation between probit-scaled ID and OOD generalization. This implies that ID accuracy is a good predictor of OOD generalization.
However, delving deeper into OOD measures is warranted for three compelling reasons.
\textbf{First}, \citet{miller2021accuracy} discuss that AoL is not universal. That is, on some datasets, ID, and OOD accuracy do not show a clear positive correlation. This point is further discussed by~\citet{wenzel2022assaying}. 
Specifically, \citet{wenzel2022assaying} suggest two patterns preventing this phenomenon: (1) underspecification (\emph{e.g.}, Camelyon17) where the same ID performance leads to different OOD behavior; (2) models do not transfer information from ID to OOD domains (\emph{e.g.}, DomainNet). That is, despite various ID performance, all models perform poorly on OOD datasets. 
\revise{Paradoxically, identifying the failure patterns itself requires labeled OOD data \citep{teney2024id}. }
\textbf{Second}, it is demanding and sophisticated to design ID test sets \citep{engstrom2020identifying}, which are expected to be unbiased and representative of distribution to effectively measure model ID accuracy. Further, it is a trade-off to split a full dataset into training, validation and test sets in terms of training and evaluation quality. 
\textbf{Third}, recent advancements in vision-language models, such as CLIP \citep{radford2021learning} and BLIP \citep{li2022blip}, have achieved remarkable zero-shot classification performance. This means that collecting a dataset for a specific task to train such models is not required.
Due to their diverse and large-scale training data, it might not be suitable to use a proxy dataset (e.g., ImageNet validation set) to reflect their performance. As shown in Figure \ref{fig:maxpred}, Vision-Language models (VLM) exhibit varying linear trends in terms of their ID and OOD accuracy compared to standard supervised models.
Furthermore, even among VLMs themselves, consistency in these linear trends is not always guaranteed.
These findings suggest that AoL alone does not suffice to accurately rank CLIP models.

\paragraph{Why Use Softmax Prediction Probability?}

Deep neural networks often exhibit a tendency to provide overly confident predictions, as evidenced by various studies \citep{ovadia2019can,minderer2021revisiting,guo2017calibration,hein2019relu}. Initially, this characteristic might raise doubts about the reliability of using Softmax Prediction as a measure of uncertainty on test data. However, existing research has shed light on its informative nature when analyzing test environments.
For instance, \citet{hendrycks2016baseline} demonstrated that the maximum Softmax prediction probability (MaxPred) for correctly classified samples tends to be higher than that of incorrectly classified or out-of-distribution (OOD) samples. This observation has paved the way for utilizing MaxPred in error/success prediction and ID/OOD detection. Building upon this insight, \citet{guillory2021predicting} and \citet{garg2022leveraging} have proposed leveraging MaxPred to estimate the accuracy of trained classifiers on test samples.

Moreover, Softmax probabilities can be efficiently computed without requiring any changes to the network architecture or additional data preprocessing. Additionally, they can be derived solely from unlabeled data. These characteristics endow prediction probability-based measures with substantial practical value as reliable indicators of OOD generalization. 
Inspired by the above discussion, this work aims to validate the feasibility of using model output-based methods (such as softmax probability) for assessing and ranking models.

\paragraph{Proof of concept.} The above works imply %
the prediction probability is likely to be effective in measuring the OOD performance of a pool of models.
Given an OOD test set (ImageNet-R) and various ImageNet models, we conduct a correlation study between MaxPred and classification accuracy. 
In Figure \ref{fig:maxpred}, we show that there is a relatively strong correlation between MaxPred and model accuracy ($\rho = 0.829$ and $\tau_w = 0.884$) on ImageNet-R. This indicates that MaxPred is feasible in ranking OOD performance. Based on this observation, we further explore more empirical prediction probability-based measures and develop a more effective measure that exploits more semantics reflected in the prediction probability.

\subsection{Exploring More Empirical Measures} \label{sec:empirical}
In addition to accuracy-on-the-line (AoL) and maximum prediction probability (MaxPred), we investigate other empirical measures as follows:

\paragraph{Average Thresholded Confidence with Maximum Confidence} (ATC-MC) \citep{garg2022leveraging}. This method is used to predict the performance of a specific trained model under distributional shift. Here, we deploy it to rank various models' performance on one particular OOD test set. It firstly identifies a threshold $t$ on $\mathcal{D}^S$ such that the number of samples with confidence score lower than $t$ is equal to the model's error. Then, the ATC on $\mathcal{D}^T$ is given by the number of points whose confidence is less than $t$. 
The formula is: $ATC = \mathbb{E}_{x \sim \mathcal{D}^T}[\mathbb{I}[{\arg \max}_{j \in \mathcal{Y}}\mathbf{P}_{:, j} < t]]$, where $\mathbb{I}[E]$ is the binary indicator of event $E$.

\paragraph{Softmax Gap} (SoftGap). MaxPred only uses the maximal value of Softmax vectors while disregarding values on other entries. Inspired by \citet{baldock2021deep}, we introduce SoftGap based on MaxPred which further considers the second-largest entry in a prediction vector. Specifically, it calculates the average difference between the largest and second-largest Softmax prediction probabilities over all samples. \revise{A high SoftGap indicates a confident prediction, while a low SoftGap suggests confusion between the two classes corresponding to the highest and second-highest probabilities.}

\subsection{Ours: Softmax Correlation (SoftmaxCorr)} \label{subsec:softmaxcorr}

\paragraph{Class-class correlation matrix} Given the prediction matrix $\bm{P} \in \mathbb{R}^{N \times K}$ predicted by $\phi_m$, a class correlation matrix $\bm{C} \in \mathbb{R}^{K \times K}$ is computed by $\bm{C} = \frac{\bm{P}^\top \bm{P}}{N}$. An entry ${C}_{i, j}$ indicates a correlation between prediction probabilities of class $i$ and class $j$ over all samples.
\revise{We define the sum of diagonals of the correlation matrix as intra-class correlation ($I_a$) and the sum of off-diagonals as inter-class correlation ($I_e$). The sum of $\bm{C}$ is $1$, which means $I_a = 1 - I_e$. Moreover, $trace(\bm{P}^T \bm{P}) = \sum_{i=1}^{N}\sum_{j=1}^{K} P_{i, j}^2 = \lVert \bm{P} \rVert ^2$, where $\lVert \cdot \rVert$ is the Frobenius norm of a matrix. Thus, we can derive $I_a = \frac{\lVert \bm{P} \rVert ^2}{N}$. According to \cite{cui2020towards}, $\lVert \bm{P} \rVert$ has strict opposite monotonicity, and the minimum of the entropy $H(\bm{P})$ and the maximum of $\lVert \bm{P} \rVert$ could be achieved at the same value. Therefore, $I_a$ reflects the certainty of predictions.
}

\paragraph{Motivation:} In domain generalization, it has been observed that class-class correlation encodes class confusion, thereby offering the potential for regularization to enhance model generalization ability \citep{chen2022reusing, jin2020minimum, li2021bi, cui2020towards}. Additionally, existing literature underscores the significance of both high prediction certainty and prediction diversity in identifying discriminative features \citep{yang2021exploiting, wang2020understanding,asano2019self}. 

\revise{Inspired by these insights, we can propose an OOD metric, which leverages a class-class correlation matrix to take into account the two characteristics:

\begin{itemize}[topsep=-8pt,itemsep=-0.5ex,partopsep=1ex,parsep=1ex,leftmargin=*]

\item Prediction certainty: the model's ability to produce confident predictions. Given the strict opposite monotonicity between $I_a$ and entropy $H(\bm{P})$, prediction certainty can be reflected by a high $I_a$ and consequently a low $I_e = 1 - I_a$ within the class correlation matrix $\bm{C}$.

\item 
Prediction diversity: relying solely on prediction confidence may be inadequate. For instance, a model may exhibit high certainty yet be biased towards a single class, making the diagonal entries of $\bm{C}$ skewed. In such cases, while prediction confidence is high, it does not necessarily indicate high model performance. Driven by this, we further evaluate prediction diversity: each class should be predicted and involved in predictions. To measure this, we expect the diagonals of $\bm{C}$ to be well-distributed across all classes instead of biased towards few classes.
\end{itemize}

Combining two characteristics, a desirable class correlation matrix has zeros on the off-diagonal entries, and values on diagonal elements match the distribution of classes. The similarity between the class correlation matrix of an evaluated model and the desirable class correlation matrix reflects the model performance. 
}
\paragraph{Implementation:} To achieve this, we propose \textbf{SoftmaxCorr} and define it as the cosine similarity between the class-class correlation matrix~$\bm{C}$ and a reference matrix~$\bm{R}$: \revise{$cos(\bm{C}, \bm{R}) = \frac{\sum_{i, j} (\bm{C} \odot \bm{R})_{i, j}}{\lVert \bm{C} \rVert \cdot \lVert \bm{R} \rVert}$, where~$\odot$ is the element-wise product between matrices.} 
The reference matrix captures the essence of the desirable class correlations: it is a diagonal matrix whose off-diagonal elements are $0$ while its diagonal entries mirror the class marginal distribution. 
To approximate this class distribution, we use the average prediction probability across test data generated by a zero-shot vision-language model (ViT-bigG/14-CLIPA). 
\revise{
The range of the SoftmaxCorr is $[0, 1]$. 
For a fixed reference matrix, the maximum is obtained 1) when a model confidently assigns probability $1$ to the predicted class and 2) the proportion of predictions for each class matches the class distribution.
The minimal value is achieved when a model is extremely biased towards one specific class with the prediction probability of one, while the estimated distribution for such a class is zero. 

}

\section{Experiments} \label{sec:experiment}

In this section, we first describe three setups including ImageNet, CIFAR-10, and WILDS.  Then, we analyze the experiment results of prediction probability-based measures on three setups. After that, {we study the impacts of class distribution estimator on predictive ability of SoftmaxCorr}. Furthermore, we study whether SoftmaxCorr can rank the performance of model checkpoints along the training trajectory. Also, we validate the effectiveness of SoftmaxCorr under the domain adaption setting. Lastly, we study the correlation between SoftmaxCorr and the accuracy of a single model on various OOD test sets.

\begin{table}[t]
    \begin{center}
     
	\setlength{\tabcolsep}{4.5pt}
	\small
	\begin{tabular}{cc | cc cc | cc cc cc}
	\toprule
\multirow{3}{*}{\textbf{Setup}}  & \multirow{3}{*}{\textbf{Dataset}} & \multicolumn{4}{c|}{\textbf{Validation Required}} & \multicolumn{6}{c}{\textbf{Validation Free}} \\
\cmidrule(lr){3-6}  \cmidrule(lr){7-12}
& & \multicolumn{2}{c}{\textbf{AoL}} & \multicolumn{2}{c|}{\textbf{ATC-MC}} & \multicolumn{2}{c}{\textbf{MaxPred}} & \multicolumn{2}{c}{\textbf{SoftGap}} & \multicolumn{2}{c}{\textbf{SoftmaxCorr}}\\
\cmidrule(lr){3-4}  \cmidrule(lr){5-6} \cmidrule(lr){7-8} \cmidrule(lr){9-10}  \cmidrule(lr){11-12} 
       &  & $\rho$ & $\tau_w$ & $\rho$ & $\tau_w$ & $\rho$ & $\tau_w$ & $\rho$ & $\tau_w$ & $\rho$ & $\tau_w$ \\
		\midrule
		\multicolumn{1}{l}{\multirow{7}{*}{\makecell{ImageNet}}}  
            & ImageNet-V2 &
            $0.954$ & $0.911$ & 
            $0.994$ & $0.961$ &
            $0.711$ & $0.597$ &
            $0.796$ & $0.644$ &
            $0.921$ & $0.758$ \\
  		& ImageNet-A &
            $0.593$ & $0.636$ & 
            $0.830$ & $0.895$ &
            $0.756$ & $0.813$ &
            $0.805$ & $0.846$ &
            $0.964$ & $0.915$ \\
		& ImageNet-R &
            $0.452$ & $0.431$ & 
            $0.950$ & $0.954$ &
            $0.829$ & $0.884$ &
            $0.902$ & $0.911$ &
            $0.951$ & $0.928$ \\
		& ImageNet-S &
            $0.505$ & $0.467$ & 
            $0.981$ & $0.959$ &
            $0.862$ & $0.864$ &
            $0.887$ & $0.871$ &
            $0.935$ & $0.909$ \\
            & ObjectNet &
            $0.619$ & $0.545$ & 
            $0.961$ & $0.821$ &
            $0.883$ & $0.849$ &
            $0.908$ & $0.865$ &
            $0.963$ & $0.895$ \\ 
            & ImageNet-Blur &
            $0.637$ & $0.576$ & 
            $0.937$ & $0.905$ &
            $0.816$ & $0.821$ &
            $0.844$ & $0.845$ &
            $0.961$ & $0.907$ \\ 
		\cmidrule(lr){2-12}
		& Average & $0.627$ & $0.594$
                        & \textcolor{blue}{$0.942$} & $\mathbf{0.916}$
                        & $0.810$ & $0.805$
                        & $0.857$ & $0.830$
                        & $\mathbf{0.949}$ & \textcolor{blue}{$0.885$}\\
	\midrule
		\multicolumn{1}{l}{\multirow{4}{*}{{CIFAR-10}}}
		& CIFAR-10.2 &
            $0.983$ & $0.949$ & 
            $0.992$ & $0.966$ &
            $0.837$ & $0.809$ &
            $0.854$ & $0.818$ &
            $0.894$ & $0.836$ \\ 
		& CINIC   &
            $0.866$ & $0.890$ & 
            $0.952$ & $0.902$ &
            $0.665$ & $0.663$ &
            $0.690$ & $0.681$ &
            $0.821$ & $0.763$ \\ 
            & CIFAR-10-Noise  &
            $0.641$ & $0.765$ & 
            $0.219$ & $0.461$ &
            $0.051$ & $0.131$ &
            $0.139$ & $0.213$ &
            $0.931$ & $0.917$ \\  
		\cmidrule(lr){2-12}
		& Average  & \textcolor{blue}{$0.830$} & $\mathbf{0.868}$
                        & $0.721$ & $0.776$
                        & $0.518$ & $0.534$
                        & $0.561$ & $0.571$
                        & $\mathbf{0.892}$ & \textcolor{blue}{$0.846$}\\
	\midrule
		\multicolumn{1}{l}{\multirow{3}{*}{{WILDS}}}
		& Camelyon17-OOD &
            $-0.021$ & $-0.072$ & 
            $-0.111$ & $-0.075$ &
            $0.192$ & $0.320$ &
            $0.192$ & $0.320$ &
            $0.420$ & $0.560$ \\ 
		& DomainNet-OOD &
            $0.350$ & $0.219$ & 
            $0.513$ & $0.254$ &
            $0.403$ & $0.274$ &
            $0.407$ & $0.258$ &
            $0.740$ & $0.680$ \\  
            
		\cmidrule(lr){2-12}
		& Average  & \textcolor{blue}{$0.165$} & \textcolor{blue}{$0.074$}
                        & $0.201$ & $0.217$
                        & $0.298$ & $0.597$
                        & $0.300$ & $0.289$
                        & $\mathbf{0.580}$ & $\mathbf{0.620}$\\
            \cmidrule(lr){1-12}
            \multicolumn{2}{c|}{Average over three setups} & 
            $0.598$ & $0.575$ &
            \textcolor{blue}{$0.747$} & \textcolor{blue}{$0.751$} &
            $0.637$ & $0.693$ &
            $0.675$ & $0.609$ &
            $\mathbf{0.864}$ & $\mathbf{0.824}$ \\
		\bottomrule
	\end{tabular}
 	\caption{{\textbf{Method comparison on ImageNet, CIFAR-10, WILDS and DomainNet}}. We compare SoftmaxCorr with four measures: accuracy-on-the-line (AoL) \citep{miller2021accuracy}, average thresholded confidence with maximum confidence (ATC-MC) \citep{garg2022leveraging}, MaxPred \citep{hendrycks2016baseline} and SoftGap~\citep{baldock2021deep}. Spearman's rank correlation ($\rho$) and weighted Kendall's correlation ($\tau_w$) are metrics. The highest correlation in each row is highlighted in \textbf{bold} and the second highest is in \textcolor{blue}{blue}. We show that our method is stable and yields the highest average correlations over three setups.
    	}\label{tab:corr}
	\end{center}
	\vspace{-10pt}
\end{table}

\subsection{Experimental Setup}
\textbf{ImageNet setup.} 
We collect 173 models publicly accessible from TIMM \citep{rw2019timm}. They are trained or fine-tuned on ImageNet \citep{deng2009imagenet} and have various architectures, training strategies and training paradigms.
We use five OOD datasets for the correlation study. Specifically, OOD datasets are: (1) ImageNet-V2 \citep{recht2019imagenet}; (2) ObjectNet \citep{barbu2019objectnet}; (3) ImageNet-S(ketch) \citep{wang2019learning}; (4) ImageNet-Blur severity $5$ \citep{hendrycks2019benchmarking}; (5) ImageNet-R(endition) \citep{hendrycks2021many}; ImageNet-R and ObjectNet contain $200$ and $113$ ImageNet classes respectively. 

\paragraph{CIFAR-10 setup.}
We collect $65$ networks trained with the scheme provided by \citet{kl2017cifar} on CIFAR-10 training set \citep{krizhevsky2009learning}. These models have different model architectures. CIFAR-10-Val(idation) is the ID test set. For OOD datasets, we use (1) CIFAR-10.2 \citep{recht2018cifar10.1} 
(2) CINIC \citep{darlow2018cinic} 
(3) CIFAR-10-Noise with severity $5$ \citep{hendrycks2019benchmarking}.
We use accuracy as the metric of model generalization.

\paragraph{WILDS setup.} We consider a classification tasks of this setup: Camelyon17 \citep{bandi2018detection}.
It is a binary classification dataset where the objective is to classify whether a slide contains a tumor issue. We use $45$ models varying in architectures and random seeds. 
ID and OOD datasets are the default ID validation set and OOD test set respectively. 
For each task, we follow the same training scheme provided by \citet{koh2021wilds} to train or fine-tune models.  

\paragraph  {Zero-shot vision language models.} In addition to models which are trained on the ID training dataset, we also consider $89$ zero-shot vision-language models, including CLIP~\citep{radford2021learning}, SigLIT~\citep{zhai2023sigmoid}, BLIP~\citep{li2022blip}, BLIP-2~\citep{li2023blip2} and Flava \citep{singh2022flava}. We use default prompt set for corresponding models. If the default prompt sets are not provided, ``\texttt{A picture of \{class\}.}'' is deployed. Unless specified, we use ViT-bigG/14-CLIPA to estimate class distribution for all setups.

\begin{figure*}[t]
    \centering
    \vspace{-5pt}
\includegraphics[width=0.85\linewidth]{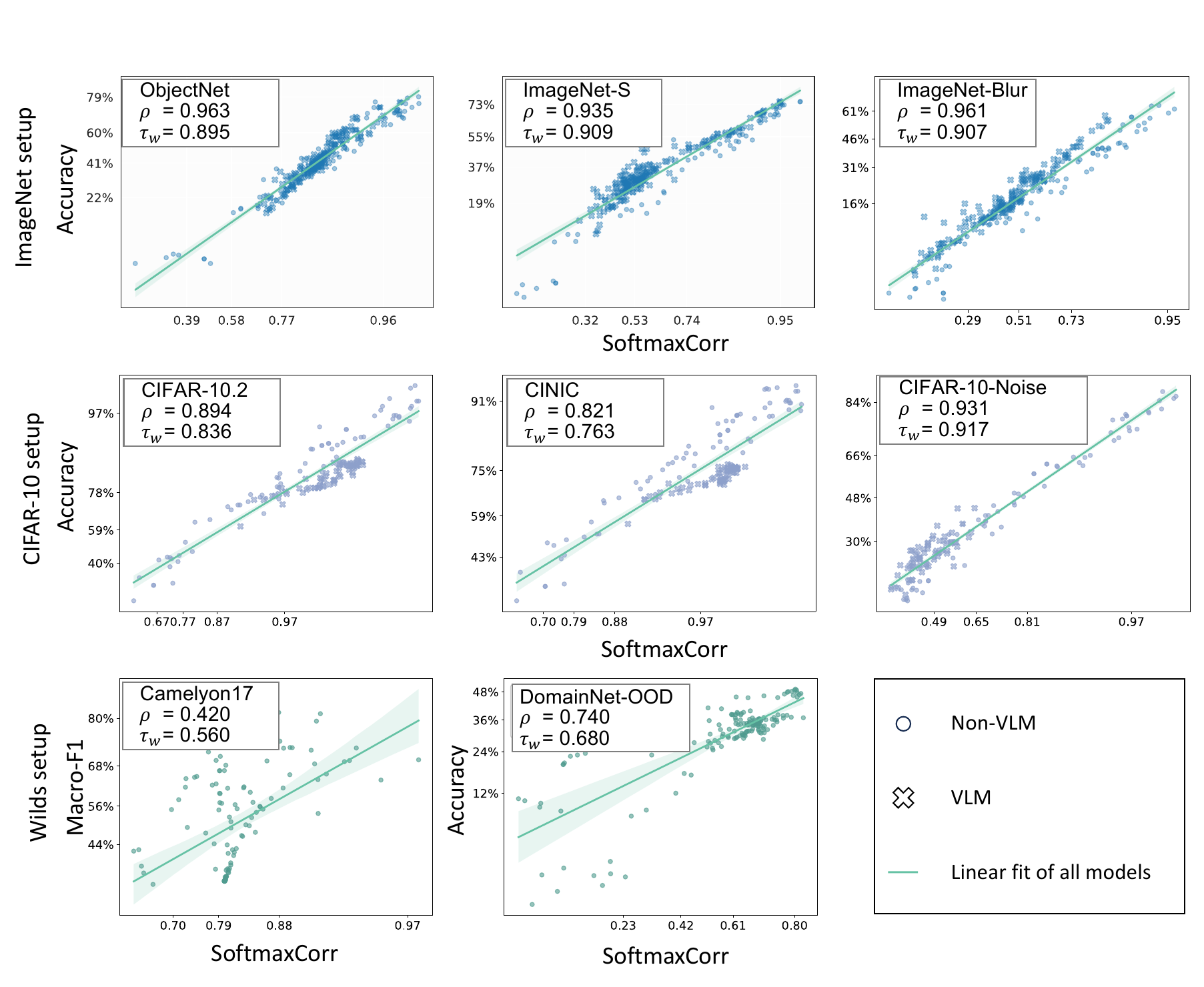}
\vspace{-10pt}
    \caption{
    \textbf{SoftmaxCorr \emph{vs.} model generalization under ImageNet, CIFAR-10 and WILDS setups}. {In each subfigure, each point denotes a model trained for the corresponding task.} For ImageNet setup, OOD test sets are ObjectNet and ImageNet-S and ImageNet-Blur.  For CIFAR-10 setup, OOD test sets are CIFAR-10.2, CINIC and CIFAR-10-Noise. 
    For WILDS, OOD test sets are Camelyon17-OOD and DomainNet-OOD.
    The $y$-axis is top-1 accuracy, top-1 accuracy and macro-F1 for the three setups, respectively. 
    Straight lines are fit with robust linear regression \citep{huber2011robust}. Axes are probit scaled as described in Section \ref{sec:formulation}. We observe that SoftmaxCorr is a reliable and effective metric. Particularly on ImageNet, SoftmaxCorr is predictive of model generalization with strong performance ($\rho > 0.92$). }
    \label{fig:softmaxcorr}
    \vspace{-10pt}
\end{figure*}

\subsection{Main Observations}

\textbf{Albeit requiring no access to validation/proxy set, SoftmaxCorr exhibits a strong correlation with model generalization.} In Figure \ref{fig:softmaxcorr} and Table \ref{tab:corr}, we observe that SoftmaxCorr is indicative of model performance under the three setups. Particularly on the ImageNet setup, SoftmaxCorr has consistently strong correlations with models' OOD performance. For example, the average Spearman’s Rank Correlation coefficient $\rho$ is $0.949$, $0.627$, \revise{$0.942$}, $0.810$, and $0.857$ for our method, AoL, ATC-MC, MaxPred and SoftGap, respectively. We also notice that AoL exhibits very strong correlation with model accuracy on ImageNet-V2 ($\rho = 0.954$), but such a correlation drops noticeably on other OOD datasets. This implies that it is insufficient to rank both supervised models and vision--language models based solely on this phenomenon.
On CIFAR-10 and WILDS, while on some test sets it does not present the strongest correlation, we still think that SoftmaxCorr is a preferred measure because it has very competitive average correlation scores. For Camelyon17, we see that all methods are less useful. We speculate this is caused by the under-specification phenomenon \citep{wenzel2022assaying} where the model relies on spurious features.

\paragraph{SoftmaxCorr gives stable correlation, while the other four measures have mixed performance on different test sets.} On CIFAR-10-Noise, we find that SoftmaxCorr correlates well with model performance ($\rho = 0.931$ and $\tau_w = 0.917$). In contrast, AoL, ATC-MC, MaxPred, SoftGap show a weaker correlation.
Although ATC-MC has a higher average weighted Kendall's correlation than SoftmaxCorr ($\rho = 0.916$ \textit{vs.} $0.885$) on ImageNet setup, it exhibits a weaker correlation on CIFAR-10 and WILDS setups.
While in some cases SoftmaxCorr does not present the highest correlation, we emphasize that it overall gives more stable and stronger correlations. Thus, we think SoftmaxCorr is generally indicative of model generalization. 

\paragraph{Compared to MaxPred and SoftGap, SoftmaxCorr better utilizes Softmax prediction probabilities.}
We use SoftGap on top of MaxPred as a simple approach to explicitly consider more entries (the second largest probability) in Softmax predictions. 
As shown in Table \ref{tab:corr}, SoftmaxCorr and SoftGap both achieve higher correlation results than MaxPred ($\rho = 0.864, 0.675$ and $0.637$, respectively). This indicates that it is helpful to analyze the distribution of softmax outputs.
Compared with MaxPred and SoftGap, the class-wise correlation considered in our SoftmaxCorr better reveals the knowledge encoded by Softmax predictions. This is supported by the higher average correlation and more stable performance of SoftmaxCorr.

\begin{table}
\begin{minipage}{.5\linewidth}
\small
    \centering
        \vspace{-8pt}
            \setlength{\tabcolsep}{2pt}
    \begin{tabular}{l c c c}
        \toprule
         \textbf{Dataset} &  \multicolumn{1}{c}{\textbf{\footnotesize{Certainty}}} & \multicolumn{1}{c}{\textbf{\footnotesize{Diversity}}}& \multicolumn{1}{c}{\textbf{\footnotesize{SoftmaxCorr}}}\\ 
        \midrule
         ImageNet-V2-A & $0.619$ & $0.551$ & $\mathbf{0.921}$ \\
         ImageNet-R & $0.851$ & $0.743$ & $\mathbf{0.951}$ \\
         ObjectNet & $0.881$ & $0.812$ & $\mathbf{0.963}$\\
         CIFAR-10.2 &  $0.881$ & $0.766$ & $\mathbf{0.894}$ \\
         CINIC & $0.643$ & $0.543$ & $\mathbf{0.821}$  \\
         DomainNet-OOD & $0.412$ & $0.139$ & $\mathbf{0.740}$  \\
        \bottomrule
    \end{tabular}
    \caption{Comparison between SoftmaxCorr and two variants (Certainty and Diversity).
    Rank correlation ($\rho$) is used as the metric. 
    } %
    \label{tab:intra}
\end{minipage}
\quad
\begin{minipage}{.45\linewidth}
\small
    \centering
            \vspace{-3pt}
      \setlength{\tabcolsep}{3pt}
        \begin{tabular}{l c c c c c}
        \toprule
         \textbf{Dataset} &  \multicolumn{1}{c}{\textbf{{\textbf{1\%}}}} & \multicolumn{1}{c}{\textbf{{\textbf{5\%}}}} & \multicolumn{1}{c}{\textbf{{\textbf{10\%}}}} & \multicolumn{1}{c}{\textbf{{\textbf{30\%}}}} & \multicolumn{1}{c}{\textbf{{\textbf{100\%}}}}\\ 
        \midrule
         ImageNet-V2 &  $0.851$ & $0.791$ & $0.817$ & $0.862$ & $0.921$\\
         ImageNet-R & $0.913$ & $0.949$ & $0.945$ & $0.951$ & $0.951$\\
         ObjectNet & $0.906$ & $0.950$ & $0.955$ & $0.963$ & $0.963$ \\
         CIFAR-10-Noise & $0.910$ & $0.933$ & $0.930$ & $0.930$ & $0.931$\\
         CINIC & $0.596$ & $0.838$ & $0.837$ & $0.832$ & $0.821$ \\
         DomainNet-OOD & $0.758$ & $0.784$ & $0.789$ & $0.789$ & $0.740$ \\
        \bottomrule
    \end{tabular}
    \vspace{-2pt}
    \caption{{\textbf{Sensitivity analysis of SoftmaxCorr on test set sizes}. We test four sampling ratios and report $\rho$ on six datasets. SoftmaxCorr is stable given a reasonable number of samples.}} %
    \label{tab:stratified}
    \vspace{-5pt}
\end{minipage}
\end{table}

\subsection{Sensitivity Analysis on Test Set Size}
We study the sensitivity of SoftmaxCorr to test set size. {Specifically, we reduce the dataset size by randomly sampling $1\%, 5\%, 10\%$ and $30\%$ of the original data. We report the averaged Spearman's correlation of three random runs on six datasets (\emph{e.g.}, ImageNet-V2, ImageNet-R, ObjectNet, CINIC, CIFAR-10-Noise and DomainNet-OOD). As shown in Table \ref{tab:stratified}, we observe that when the number of test data is very small ($1\%$), the correlation of SoftmaxCorr drops. When the dataset size increases ($\ge 10\%$), SoftmaxCorr exhibits a stable and high correlation with model performance. This suggests that SoftmaxCorr requires a reasonable number of samples to capture model generalization.
}

\begin{table}[t]
    \begin{center}
     
	\setlength{\tabcolsep}{4.5pt}
	\small
    \begin{tabular}{l c c c c c c}
        \toprule
         \textbf{Method} &  \multicolumn{1}{c}{\textbf{{ImageNet-V2-A}}} & \multicolumn{1}{c}{\textbf{{ImageNet-R}}}& \multicolumn{1}{c}{\textbf{{ObjectNet}}} & 
         \multicolumn{1}{c}{\textbf{{CIFAR-10.2}}} & 
         \multicolumn{1}{c}{\textbf{{CINIC}}} & 
         \multicolumn{1}{c}{\textbf{{DomainNet-OOD}}}\\ 
        \midrule
         Disagreement & $-0.379$ & $0.746$ & $0.021$ & $0.264$ & $0.314$ & $-0.083$\\
         SoftmaxCorr & $0.921$ & $0.951$ & $0.963$ & $0.894$ & $0.821$ & $0.740$\\
         \bottomrule
 	\end{tabular}
        \caption{ Comparison between SoftmaxCorr and Disagreement. Rank correlation ($\rho$) is used as the metric.
    	}\label{tab:disagree}
	\end{center}
	\vspace{-10pt}
\end{table}

\subsection{Impacts of Class Distribution Estimator}
\revise{In previous experiments, we solely use a vision-language model as the reference model to estimate class distribution. Another baseline usage of such a reference model is to measure the disagreements between the predictions of evaluated models and the reference model (Disagreement). In Table \ref{tab:disagree}, we compare Spearman's rank correlation of Disagreement and SoftmaxCorr. We observe that SoftmaxCorr consistently gives a stronger correlation than Disagreement. Disagreement explicitly measures the similarity between the evaluated models and the reference model, so the performance of the reference model significantly influences the predictive ability of Disagreement. 
In contrast, SoftmaxCorr only uses it for estimating the empirical class distribution, which constitutes the diagonal elements of the reference matrix $\bm{R}$, with all off-diagonal elements set to zero.
}

\begin{figure*}[t]
    \centering
    \includegraphics[width=0.85\linewidth]{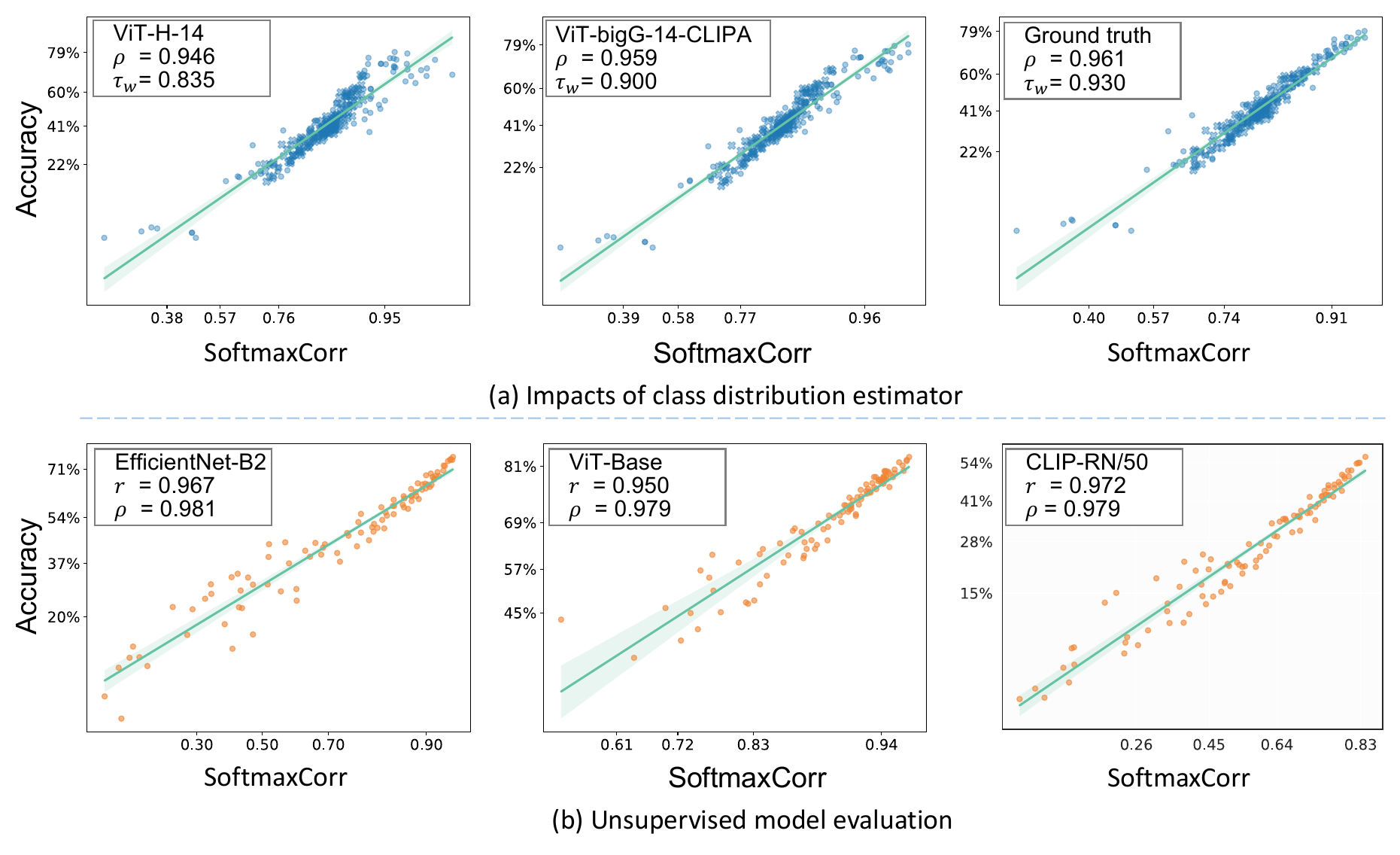}
    \vspace{-10pt}
    \caption{\textbf{(a) Impacts of class distribution estimator}, we use three estimators: ViT-H-14, ViT-bigG-14-CLIPA and the ground truth. We find SoftmaxCorr is fairly stable. \textbf{(b) SoftmaxCorr \textit{v.s.} accuracy on ImageNet-C benchmark}. {In every subfigure, each dot indicates a dataset of ImageNet-C.}. We see strong correlations between SoftmaxCorr and OOD accuracy on various test set.}
    \label{fig:analysis}
    \vspace{-10pt}
\end{figure*}

To study the influence caused by the estimator, we further use a less accurate zero-shot vision-language model (ViT-H-14) and an ideal estimator for calculating the marginal class distribution. We evaluate them on ObjectNet, because it has imbalanced class distribution. In Figure \ref{fig:analysis}, we observe that with a moderately accurate ViT-H-14, SoftmaxCorr remains predictive, and the reference estimator further enhances SoftmaxCorr. Note that, we only use the average of softmax prediction probability to estimate class weights. Some better-designed algorithms \citep{lipton2018detecting, garg2020unified, sun2022prior} may improve SoftmaxCorr's stability, and we leave it as future work.

\subsection{Different Characterizations of Correlation Matrix} 
To investigate the importance of prediction diversity and certainty for predicting OOD generalization, we compare SoftmaxCorr with two variants: (1) the sum of diagonal entries in the class correlation matrix (Certainty). It measures whether the values in diagonal entries are large, indicating prediction certainty;
{(2) the Euclidean distance between diagonal elements in the class correlation matrix and estimated class distribution (Diversity). It measures whether models make diverse predictions whose distribution matches the estimated distribution of each class.}
In Table \ref{tab:intra}, we present Spearman's rank correlation of three methods on six datasets from three setups. Both variants give weaker correlation strength than SoftmaxCorr. Specifically, SoftmaxCorr is more predictive of OOD generalization than Certainty and Diversity ($\rho = 0.821$ \textit{vs.} $0.643$ \textit{vs.} {$0.543$}) on CINIC. {This indicates that it is important to measure both prediction diversity and certainty for OOD generalization assessment.}

\subsection{Effectiveness of SoftmaxCorr for Domain Adaptation}

{On ImageNet, CIFAR and WILDS setups, all models are trained by standard empirical risk minimization and do not use the unlabeled OOD samples from training. In some scenarios, domain adaptation (DA) algorithms are employed for learning target-adaptive models with additional unlabeled OOD samples \citep{kouw2019review, zhou2022domain}. To explore whether SoftmaxCorr is still effective in assessing the generalization of these models, we conduct a correlation study under DomainNet setup \citep{peng2019moment, sagawa2021extending}. The models are trained by 9 different DA algorithms (\emph{e.g.}, DeepCORAL \citep{sun2016deep}, DANN \citep{dann}). In Table~\ref{tab:corr}, we observe that SoftmaxCorr performs reasonably on DomainNet-OOD. We also notice that ID accuracy-based methods (AoL and ATC-MC) become less useful. DA algorithms likely focus on improving OOD performance, while ID accuracy may not be enhanced accordingly.}

\subsection{SoftmaxCorr Reflects a Model’s Generalization on Various OOD Test Sets}

We investigate how a given trained model generalizes to various OOD datasets. In detail, we evaluate a single model on all test sets of ImageNet-C benchmark \citep{hendrycks2019benchmarking} and conduct a correlation study between accuracy and SoftmaxCorr. We additionally use Pearson's correlation ($r$) to measure the overall linear trend. This coefficient varies in $[-1, 1]$. A value closer to $-1/1$ indicates better negative/positive linearity and $0$ means no correlation. We use ImageNet networks: EfficientNet-B2 \citep{tan2019efficientnet}, ViT-Base \citep{dosovitskiy2020image} and CLIP-RN/50 \citep{radford2021learning}. Figure \ref{fig:analysis} shows a strong linear relationship and high-rank correlation ($r > 0.95$ and $\rho > 0.97$). It indicates that with linear regression, SoftmaxCorr can also help estimate the accuracy of a given model on various test sets.

\begin{figure*}[t]
    \centering
    \includegraphics[width=\linewidth]{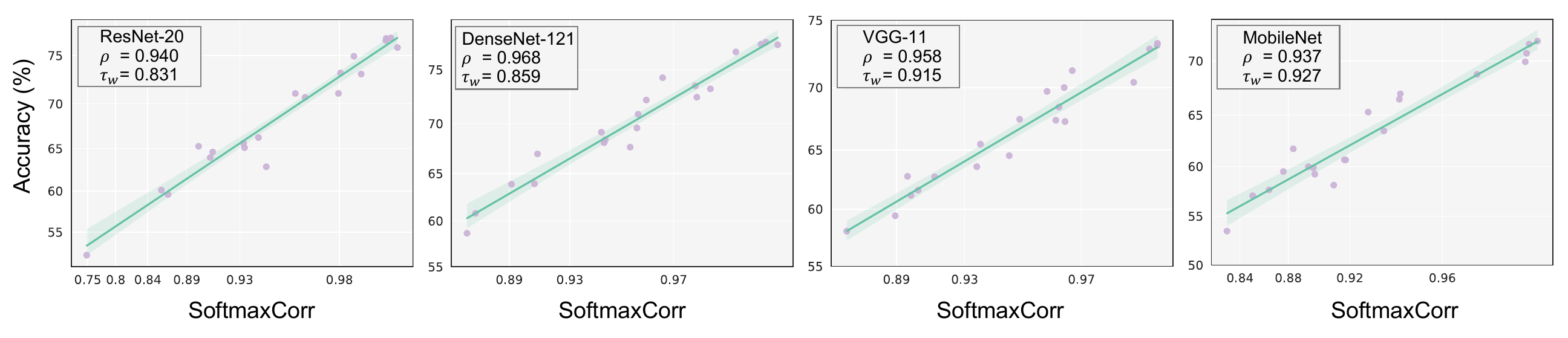}
    \vspace{-20pt}
    \caption{
   \textbf{Correlation analysis: SoftmaxCorr and accuracy on CINIC}. Each point represents a checkpoint. We consider CIFAR-10 models: ResNet-20, DenseNet-121, VGG-11 and MobileNet. 
    Axes are probit scaled as in Section \ref{sec:formulation}. {In each subfigure, every point means a checkpoint of the model along the training process.}
    For four models, we see strong correlations ($\rho > 0.93 $). This suggests that SoftmaxCorr is helpful in assessing checkpoints along the training process.
    }
    \vspace{-10pt}
    \label{fig:trajectory}
\end{figure*}

\subsection{Evaluation Along Training Trajectory}
In previous sections, SoftmaxCorr is used to measure the performance of models varying in different architectures and training strategies. In practice, we are sometimes interested in evaluating the models at different training checkpoints. %
Hence, we analyze whether SoftmaxCorr is helpful in this case. We collect prediction probabilities on {CINIC} every $10$ epochs along the training process of ResNet-20, DenseNet-121 \citep{huang2017densely}, VGG11 \citep{simonyan2014very} and MobileNet \citep{howard2017mobilenets} trained on CIFAR-10. In Fig. \ref{fig:trajectory}, we observe SoftmaxCorr has a high-rank correlation ($\rho > 0.93$) with model performance. This means we can potentially apply SoftmaxCorr to assay model generalization along the training process.

\section{Discussion and Potential Directions}

\paragraph{Discussion on imbalanced test sets.} 
SoftmaxCorr is defined as cosine similarity between correlation matrix $\bm{C}$ and a reference matrix~$\bm{R}_K$. The reference matrix is a diagonal matrix and each entry represents the corresponding class distribution. To derive class distribution, a zero-shot vision-language model is used for efficient deployment without training and shows strong robustness towards distribution shifts. Consequently, the predictive ability of SoftmaxCorr correlates to the deployed vision-language model and the method used to estimate the distribution. Hence, it would be beneficial to use advanced label shift estimation techniques \citep{lipton2018detecting, garg2020unified, sun2022prior} and employ more performant vision-language models, and we leave it as future work. 

\paragraph{Potential OOD measures.} This work proposes SoftmaxCorr to use class-wise relationships encoded by Softmax prediction probabilities. Here, we discuss other potential ways. \textbf{First}, SoftGap computes the difference between the largest and second-largest prediction probabilities. 
We show that SoftGap exhibits a stronger correlation with performance compared to MaxPred. 
It would be interesting to improve SoftGap by utilizing more probabilities (\emph{e.g.}, top five probabilities).
{\textbf{Second}, for a perfectly calibrated model, its MaxPred over a test set corresponds to its accuracy. Yet, calibration methods seldom exhibit desired performance under distribution shift \citep{ovadia2019can}. That said, it would be promising to study post-hoc calibration methods for OOD datasets, which benefits MaxPred for assessing model generalization.
\textbf{Last}, this work focuses on Softmax prediction probability. We tested our method based on logits but no obvious correlation is exhibited. This may be because the logits of different models vary in significantly different ranges. We also think that studying other model statistics (\emph{e.g.}, weights and feature representations) would be interesting. 

\section{Conclusion}
This paper studies an important problem of assaying and ranking model generalization under distribution shifts. To this end, we explore the use of Softmax prediction probability for developing OOD measures. We start by identifying the demand for OOD measures beyond accuracy-on-the-line and finding that maximum Softmax prediction probability is to some extent useful to measure the OOD performance. We then propose Softmax Correlation (SoftmaxCorr) which leverages class confusion encoded by the class-class correlation matrix and thus better reflects the overall quality of the classifier predictions. To validate the usefulness of SoftmaxCorr, we compare it with four other empirical measures across $11$ datasets under ImageNet, CIFAR-10 and WILDS setups. We observe SoftmaxCorr generally presents a stable and high correlation with model performance on different OOD datasets.
This paper establishes some baseline usage of Softmax prediction probability and a specific improvement, and more investigation will be made in the future.

\subsubsection*{Acknowledgements}
{We sincerely thank all the anonymous reviewers and area chairs for their constructive comments and valuable suggestions, which have greatly helped in enhancing this paper.}

\bibliography{main}

\begin{thebibliography}{109}
\providecommand{\natexlab}[1]{#1}
\providecommand{\url}[1]{\texttt{#1}}
\expandafter\ifx\csname urlstyle\endcsname\relax
  \providecommand{\doi}[1]{doi: #1}\else
  \providecommand{\doi}{doi: \begingroup \urlstyle{rm}\Url}\fi

\bibitem[Acuna et~al.(2021)Acuna, Zhang, Law, and Fidler]{acuna2021f}
David Acuna, Guojun Zhang, Marc~T Law, and Sanja Fidler.
\newblock f-domain adversarial learning: Theory and algorithms.
\newblock In \emph{International Conference on Machine Learning}, pp.\  66--75,
  2021.

\bibitem[Alhamoud et~al.(2022)Alhamoud, Hammoud, Alfarra, and
  Ghanem]{alhamoud2022generalizability}
Kumail Alhamoud, Hasan Abed Al~Kader Hammoud, Motasem Alfarra, and Bernard
  Ghanem.
\newblock Generalizability of adversarial robustness under distribution shifts.
\newblock \emph{Transaction on Machine Learning Research}, 2022.

\bibitem[Arora et~al.(2018)Arora, Ge, Neyshabur, and Zhang]{arora2018stronger}
Sanjeev Arora, Rong Ge, Behnam Neyshabur, and Yi~Zhang.
\newblock Stronger generalization bounds for deep nets via a compression
  approach.
\newblock In \emph{International Conference on Machine Learning}, pp.\
  254--263, 2018.

\bibitem[Asano et~al.(2019)Asano, Rupprecht, and Vedaldi]{asano2019self}
Yuki~Markus Asano, Christian Rupprecht, and Andrea Vedaldi.
\newblock Self-labelling via simultaneous clustering and representation
  learning.
\newblock In \emph{International Conference on Learning Representations}, 2019.

\bibitem[Baek et~al.(2022)Baek, Jiang, Raghunathan, and
  Kolter]{baek2022agreement}
Christina Baek, Yiding Jiang, Aditi Raghunathan, and Zico Kolter.
\newblock Agreement-on-the-line: Predicting the performance of neural networks
  under distribution shift.
\newblock In \emph{Advances in Neural Information Processing Systems}, 2022.

\bibitem[Baldock et~al.(2021)Baldock, Maennel, and Neyshabur]{baldock2021deep}
Robert Baldock, Hartmut Maennel, and Behnam Neyshabur.
\newblock Deep learning through the lens of example difficulty.
\newblock In \emph{Advances in Neural Information Processing Systems},
  volume~34, pp.\  10876--10889, 2021.

\bibitem[Bandi et~al.(2018)Bandi, Geessink, Manson, Van~Dijk, Balkenhol,
  Hermsen, Bejnordi, Lee, Paeng, Zhong, et~al.]{bandi2018detection}
Peter Bandi, Oscar Geessink, Quirine Manson, Marcory Van~Dijk, Maschenka
  Balkenhol, Meyke Hermsen, Babak~Ehteshami Bejnordi, Byungjae Lee, Kyunghyun
  Paeng, Aoxiao Zhong, et~al.
\newblock From detection of individual metastases to classification of lymph
  node status at the patient level: the camelyon17 challenge.
\newblock \emph{IEEE Transactions on Medical Imaging}, 2018.

\bibitem[Barbu et~al.(2019)Barbu, Mayo, Alverio, Luo, Wang, Gutfreund,
  Tenenbaum, and Katz]{barbu2019objectnet}
Andrei Barbu, David Mayo, Julian Alverio, William Luo, Christopher Wang, Dan
  Gutfreund, Josh Tenenbaum, and Boris Katz.
\newblock Objectnet: A large-scale bias-controlled dataset for pushing the
  limits of object recognition models.
\newblock In \emph{Advances in Neural Information Processing Systems}, 2019.

\bibitem[Bartlett et~al.(2017)Bartlett, Foster, and
  Telgarsky]{bartlett2017spectrally}
Peter~L Bartlett, Dylan~J Foster, and Matus~J Telgarsky.
\newblock Spectrally-normalized margin bounds for neural networks.
\newblock In \emph{Advances in Neural Information Processing Systems}, 2017.

\bibitem[Ben-David et~al.(2006)Ben-David, Blitzer, Crammer, and
  Pereira]{ben2006analysis}
Shai Ben-David, John Blitzer, Koby Crammer, and Fernando Pereira.
\newblock Analysis of representations for domain adaptation.
\newblock In \emph{Advances in Neural Information Processing Systems}, 2006.

\bibitem[Blitzer et~al.(2007)Blitzer, Crammer, Kulesza, Pereira, and
  Wortman]{blitzer2007learning}
John Blitzer, Koby Crammer, Alex Kulesza, Fernando Pereira, and Jennifer
  Wortman.
\newblock Learning bounds for domain adaptation.
\newblock In \emph{Advances in neural information processing systems}, 2007.

\bibitem[Chen et~al.(2022)Chen, Chen, Wei, Jin, Tan, Jin, and
  Chen]{chen2022reusing}
Lin Chen, Huaian Chen, Zhixiang Wei, Xin Jin, Xiao Tan, Yi~Jin, and Enhong
  Chen.
\newblock Reusing the task-specific classifier as a discriminator:
  Discriminator-free adversarial domain adaptation.
\newblock In \emph{Proceedings of the IEEE/CVF Conference on Computer Vision
  and Pattern Recognition}, 2022.

\bibitem[Chuang et~al.(2021)Chuang, Mroueh, Greenewald, Torralba, and
  Jegelka]{chuang2021measuring}
Ching-Yao Chuang, Youssef Mroueh, Kristjan Greenewald, Antonio Torralba, and
  Stefanie Jegelka.
\newblock Measuring generalization with optimal transport.
\newblock In \emph{Advances in Neural Information Processing Systems}, pp.\
  8294--8306, 2021.

\bibitem[Cui et~al.(2020)Cui, Wang, Zhuo, Li, Huang, and Tian]{cui2020towards}
Shuhao Cui, Shuhui Wang, Junbao Zhuo, Liang Li, Qingming Huang, and Qi~Tian.
\newblock Towards discriminability and diversity: Batch nuclear-norm
  maximization under label insufficient situations.
\newblock In \emph{Proceedings of the IEEE/CVF Conference on Computer Vision
  and Pattern Recognition}, pp.\  3941--3950, 2020.

\bibitem[Darlow et~al.(2018)Darlow, Crowley, Antoniou, and
  Storkey]{darlow2018cinic}
Luke~N Darlow, Elliot~J Crowley, Antreas Antoniou, and Amos~J Storkey.
\newblock Cinic-10 is not imagenet or cifar-10.
\newblock \emph{arXiv preprint arXiv:1810.03505}, 2018.

\bibitem[Deng et~al.(2022)Deng, Li, Xiong, Chen, and Hooi]{Deng2022trust}
Ailin Deng, Shen Li, Miao Xiong, Zhirui Chen, and Bryan Hooi.
\newblock Trust, but verify: Using self-supervised probing to improve
  trustworthiness.
\newblock In \emph{European Conference on Computer Vision}, 2022.

\bibitem[Deng et~al.(2009)Deng, Dong, Socher, Li, Li, and
  Fei-Fei]{deng2009imagenet}
Jia Deng, Wei Dong, Richard Socher, Li-Jia Li, Kai Li, and Li~Fei-Fei.
\newblock Imagenet: A large-scale hierarchical image database.
\newblock In \emph{Proceedings of the IEEE Conference on Computer Vision and
  Pattern Recognition}, pp.\  248--255, 2009.

\bibitem[Deng \& Zheng(2021)Deng and Zheng]{deng2021labels}
Weijian Deng and Liang Zheng.
\newblock Are labels always necessary for classifier accuracy evaluation?
\newblock In \emph{Proceedings of the IEEE/CVF Conference on Computer Vision
  and Pattern Recognition}, pp.\  15069--15078, 2021.

\bibitem[Deng et~al.(2023)Deng, Suh, Gould, and Zheng]{deng2023confidence}
Weijian Deng, Yumin Suh, Stephen Gould, and Liang Zheng.
\newblock Confidence and dispersity speak: Characterising prediction matrix for
  unsupervised accuracy estimation.
\newblock In \emph{International Conference on Machine Learning}, 2023.

\bibitem[Dinh et~al.(2017)Dinh, Pascanu, Bengio, and Bengio]{dinh2017sharp}
Laurent Dinh, Razvan Pascanu, Samy Bengio, and Yoshua Bengio.
\newblock Sharp minima can generalize for deep nets.
\newblock In \emph{International Conference on Machine Learning}, pp.\
  1019--1028, 2017.

\bibitem[Djolonga et~al.(2021)Djolonga, Yung, Tschannen, Romijnders, Beyer,
  Kolesnikov, Puigcerver, Minderer, D'Amour, Moldovan,
  et~al.]{djolonga2021robustness}
Josip Djolonga, Jessica Yung, Michael Tschannen, Rob Romijnders, Lucas Beyer,
  Alexander Kolesnikov, Joan Puigcerver, Matthias Minderer, Alexander D'Amour,
  Dan Moldovan, et~al.
\newblock On robustness and transferability of convolutional neural networks.
\newblock In \emph{Proceedings of the IEEE Conference on Computer Vision and
  Pattern Recognition}, pp.\  16458--16468, 2021.

\bibitem[Dosovitskiy et~al.(2020)Dosovitskiy, Beyer, Kolesnikov, Weissenborn,
  Zhai, Unterthiner, Dehghani, Minderer, Heigold, Gelly,
  et~al.]{dosovitskiy2020image}
Alexey Dosovitskiy, Lucas Beyer, Alexander Kolesnikov, Dirk Weissenborn,
  Xiaohua Zhai, Thomas Unterthiner, Mostafa Dehghani, Matthias Minderer, Georg
  Heigold, Sylvain Gelly, et~al.
\newblock An image is worth 16x16 words: Transformers for image recognition at
  scale.
\newblock In \emph{International Conference on Learning Representations}, 2020.

\bibitem[Dziugaite \& Roy(2017)Dziugaite and Roy]{dziugaite2017computing}
Gintare~Karolina Dziugaite and Daniel~M Roy.
\newblock Computing nonvacuous generalization bounds for deep (stochastic)
  neural networks with many more parameters than training data.
\newblock In \emph{Proceedings of the Thirty-Third Conference on Uncertainty in
  Artificial Intelligence}, 2017.

\bibitem[Dziugaite et~al.(2020)Dziugaite, Drouin, Neal, Rajkumar, Caballero,
  Wang, Mitliagkas, and Roy]{dziugaite2020search}
Gintare~Karolina Dziugaite, Alexandre Drouin, Brady Neal, Nitarshan Rajkumar,
  Ethan Caballero, Linbo Wang, Ioannis Mitliagkas, and Daniel~M Roy.
\newblock In search of robust measures of generalization.
\newblock In \emph{Advances in Neural Information Processing Systems}, pp.\
  11723--11733, 2020.

\bibitem[Engstrom et~al.(2020)Engstrom, Ilyas, Santurkar, Tsipras, Steinhardt,
  and Madry]{engstrom2020identifying}
Logan Engstrom, Andrew Ilyas, Shibani Santurkar, Dimitris Tsipras, Jacob
  Steinhardt, and Aleksander Madry.
\newblock Identifying statistical bias in dataset replication.
\newblock In \emph{International Conference on Machine Learning}, pp.\
  2922--2932. PMLR, 2020.

\bibitem[Foret et~al.(2020)Foret, Kleiner, Mobahi, and
  Neyshabur]{foret2020sharpness}
Pierre Foret, Ariel Kleiner, Hossein Mobahi, and Behnam Neyshabur.
\newblock Sharpness-aware minimization for efficiently improving
  generalization.
\newblock In \emph{International Conference on Learning Representations}, 2020.

\bibitem[Ganin et~al.(2016)Ganin, Ajakan, Larochelle, Marchand, and
  Lempitsky]{dann}
Ustinova Ganin, Germain Ajakan, Laviolette Larochelle, Marchand, and Lempitsky.
\newblock Domain-adversarial training of neural networks.
\newblock In \emph{Journal of Machine Learning Research}, 2016.

\bibitem[Garg et~al.(2020)Garg, Wu, Balakrishnan, and Lipton]{garg2020unified}
Saurabh Garg, Yifan Wu, Sivaraman Balakrishnan, and Zachary Lipton.
\newblock A unified view of label shift estimation.
\newblock In \emph{Advances in Neural Information Processing Systems}, pp.\
  3290--3300, 2020.

\bibitem[Garg et~al.(2022)Garg, Balakrishnan, Lipton, Neyshabur, and
  Sedghi]{garg2022leveraging}
Saurabh Garg, Sivaraman Balakrishnan, Zachary~C Lipton, Behnam Neyshabur, and
  Hanie Sedghi.
\newblock Leveraging unlabeled data to predict out-of-distribution performance.
\newblock In \emph{International Conference on Learning Representations}, 2022.

\bibitem[Geirhos et~al.(2021)Geirhos, Narayanappa, Mitzkus, Thieringer, Bethge,
  Wichmann, and Brendel]{geirhos2021partial}
Robert Geirhos, Kantharaju Narayanappa, Benjamin Mitzkus, Tizian Thieringer,
  Matthias Bethge, Felix~A Wichmann, and Wieland Brendel.
\newblock Partial success in closing the gap between human and machine vision.
\newblock In \emph{Advances in Neural Information Processing Systems}, 2021.

\bibitem[Guillory et~al.(2021)Guillory, Shankar, Ebrahimi, Darrell, and
  Schmidt]{guillory2021predicting}
Devin Guillory, Vaishaal Shankar, Sayna Ebrahimi, Trevor Darrell, and Ludwig
  Schmidt.
\newblock Predicting with confidence on unseen distributions.
\newblock In \emph{Proceedings of the IEEE/CVF International Conference on
  Computer Vision}, pp.\  1134--1144, 2021.

\bibitem[Gulrajani \& Lopez-Paz(2021)Gulrajani and
  Lopez-Paz]{gulrajani2020search}
Ishaan Gulrajani and David Lopez-Paz.
\newblock In search of lost domain generalization.
\newblock In \emph{International Conference on Learning Representations}, 2021.

\bibitem[Guo et~al.(2017)Guo, Pleiss, Sun, and Weinberger]{guo2017calibration}
Chuan Guo, Geoff Pleiss, Yu~Sun, and Kilian~Q Weinberger.
\newblock On calibration of modern neural networks.
\newblock In \emph{International Conference on Machine Learning}, pp.\
  1321--1330, 2017.

\bibitem[He et~al.(2022)He, Chen, Xie, Li, Doll{\'a}r, and
  Girshick]{he2022masked}
Kaiming He, Xinlei Chen, Saining Xie, Yanghao Li, Piotr Doll{\'a}r, and Ross
  Girshick.
\newblock Masked autoencoders are scalable vision learners.
\newblock In \emph{Proceedings of the IEEE/CVF Conference on Computer Vision
  and Pattern Recognition}, 2022.

\bibitem[Hein et~al.(2019)Hein, Andriushchenko, and Bitterwolf]{hein2019relu}
Matthias Hein, Maksym Andriushchenko, and Julian Bitterwolf.
\newblock Why relu networks yield high-confidence predictions far away from the
  training data and how to mitigate the problem.
\newblock In \emph{Proceedings of the IEEE/CVF Conference on Computer Vision
  and Pattern Recognition}, pp.\  41--50, 2019.

\bibitem[Hendrycks \& Dietterich(2019)Hendrycks and
  Dietterich]{hendrycks2019benchmarking}
Dan Hendrycks and Thomas Dietterich.
\newblock Benchmarking neural network robustness to common corruptions and
  perturbations.
\newblock \emph{International Conference on Learning Representations}, 2019.

\bibitem[Hendrycks \& Gimpel(2016)Hendrycks and Gimpel]{hendrycks2016baseline}
Dan Hendrycks and Kevin Gimpel.
\newblock A baseline for detecting misclassified and out-of-distribution
  examples in neural networks.
\newblock In \emph{International Conference on Learning Representations}, 2016.

\bibitem[Hendrycks et~al.(2021{\natexlab{a}})Hendrycks, Basart, Mu, Kadavath,
  Wang, Dorundo, Desai, Zhu, Parajuli, Guo, et~al.]{hendrycks2021many}
Dan Hendrycks, Steven Basart, Norman Mu, Saurav Kadavath, Frank Wang, Evan
  Dorundo, Rahul Desai, Tyler Zhu, Samyak Parajuli, Mike Guo, et~al.
\newblock The many faces of robustness: A critical analysis of
  out-of-distribution generalization.
\newblock In \emph{Proceedings of the IEEE/CVF International Conference on
  Computer Vision}, pp.\  8340--8349, 2021{\natexlab{a}}.

\bibitem[Hendrycks et~al.(2021{\natexlab{b}})Hendrycks, Zhao, Basart,
  Steinhardt, and Song]{hendrycks2021natural}
Dan Hendrycks, Kevin Zhao, Steven Basart, Jacob Steinhardt, and Dawn Song.
\newblock Natural adversarial examples.
\newblock In \emph{Proceedings of the IEEE/CVF Conference on Computer Vision
  and Pattern Recognition}, pp.\  15262--15271, 2021{\natexlab{b}}.

\bibitem[Hoffman et~al.(2018)Hoffman, Mohri, and Zhang]{hoffman2018algorithms}
Judy Hoffman, Mehryar Mohri, and Ningshan Zhang.
\newblock Algorithms and theory for multiple-source adaptation.
\newblock In \emph{Advances in neural information processing systems}, 2018.

\bibitem[Howard et~al.(2017)Howard, Zhu, Chen, Kalenichenko, Wang, Weyand,
  Andreetto, and Adam]{howard2017mobilenets}
Andrew~G Howard, Menglong Zhu, Bo~Chen, Dmitry Kalenichenko, Weijun Wang,
  Tobias Weyand, Marco Andreetto, and Hartwig Adam.
\newblock Mobilenets: Efficient convolutional neural networks for mobile vision
  applications.
\newblock \emph{arXiv preprint arXiv:1704.04861}, 2017.

\bibitem[Huang et~al.(2017)Huang, Liu, Van Der~Maaten, and
  Weinberger]{huang2017densely}
Gao Huang, Zhuang Liu, Laurens Van Der~Maaten, and Kilian~Q Weinberger.
\newblock Densely connected convolutional networks.
\newblock In \emph{Proceedings of the IEEE Conference on Computer Vision and
  Pattern Recognition}, pp.\  4700--4708, 2017.

\bibitem[Huang et~al.(2022)Huang, Xia, Shen, Han, Gong, Gong, and
  Liu]{huang2022harnessing}
Zhuo Huang, Xiaobo Xia, Li~Shen, Bo~Han, Mingming Gong, Chen Gong, and
  Tongliang Liu.
\newblock Harnessing out-of-distribution examples via augmenting content and
  style.
\newblock In \emph{International Conference on Learning Representations}, 2022.

\bibitem[Huang et~al.(2023)Huang, Li, Shen, Yu, Gong, Han, and
  Liu]{huang2023winning}
Zhuo Huang, Muyang Li, Li~Shen, Jun Yu, Chen Gong, Bo~Han, and Tongliang Liu.
\newblock Winning prize comes from losing tickets: Improve invariant learning
  by exploring variant parameters for out-of-distribution generalization.
\newblock \emph{arXiv preprint arXiv:2310.16391}, 2023.

\bibitem[Huber(2011)]{huber2011robust}
Peter~J Huber.
\newblock Robust statistics.
\newblock In \emph{International Encyclopedia of Statistical Science}, pp.\
  1248--1251. Springer, 2011.

\bibitem[Ilharco et~al.(2021)Ilharco, Wortsman, Wightman, Gordon, Carlini,
  Taori, Dave, Shankar, Namkoong, Miller, Hajishirzi, Farhadi, and
  Schmidt]{ilharco_gabriel_2021_5143773}
Gabriel Ilharco, Mitchell Wortsman, Ross Wightman, Cade Gordon, Nicholas
  Carlini, Rohan Taori, Achal Dave, Vaishaal Shankar, Hongseok Namkoong, John
  Miller, Hannaneh Hajishirzi, Ali Farhadi, and Ludwig Schmidt.
\newblock Openclip, July 2021.
\newblock URL \url{https://doi.org/10.5281/zenodo.5143773}.
\newblock If you use this software, please cite it as below.

\bibitem[Jiang et~al.(2020{\natexlab{a}})Jiang, Foret, Yak, Roy, Mobahi,
  Dziugaite, Bengio, Gunasekar, Guyon, and Neyshabur]{jiang2020neurips}
Yiding Jiang, Pierre Foret, Scott Yak, Daniel~M Roy, Hossein Mobahi,
  Gintare~Karolina Dziugaite, Samy Bengio, Suriya Gunasekar, Isabelle Guyon,
  and Behnam Neyshabur.
\newblock Neurips 2020 competition: Predicting generalization in deep learning.
\newblock \emph{arXiv preprint arXiv:2012.07976}, 2020{\natexlab{a}}.

\bibitem[Jiang et~al.(2020{\natexlab{b}})Jiang, Neyshabur, Mobahi, Krishnan,
  and Bengio]{jiang2019fantastic}
Yiding Jiang, Behnam Neyshabur, Hossein Mobahi, Dilip Krishnan, and Samy
  Bengio.
\newblock Fantastic generalization measures and where to find them.
\newblock In \emph{International Conference on Learning Representations},
  2020{\natexlab{b}}.

\bibitem[Jin et~al.(2020)Jin, Wang, Long, and Wang]{jin2020minimum}
Ying Jin, Ximei Wang, Mingsheng Long, and Jianmin Wang.
\newblock Minimum class confusion for versatile domain adaptation.
\newblock In \emph{European Conference on Computer Vision}, pp.\  464--480,
  2020.

\bibitem[Kendall(1948)]{kendall1948rank}
Maurice~George Kendall.
\newblock \emph{Rank correlation methods.}
\newblock Griffin, 1948.

\bibitem[Keskar et~al.(2017)Keskar, Mudigere, Nocedal, Smelyanskiy, and
  Tang]{keskar2016large}
Nitish~Shirish Keskar, Dheevatsa Mudigere, Jorge Nocedal, Mikhail Smelyanskiy,
  and Ping Tak~Peter Tang.
\newblock On large-batch training for deep learning: Generalization gap and
  sharp minima.
\newblock In \emph{International Conference on Learning Representations}, 2017.

\bibitem[Kirsch \& Gal(2022)Kirsch and Gal]{kirsch2022note}
Andreas Kirsch and Yarin Gal.
\newblock A note on" assessing generalization of sgd via disagreement".
\newblock \emph{Transaction on Machine Learning Research}, 2022.

\bibitem[Koh et~al.(2021)Koh, Sagawa, Marklund, Xie, Zhang, Balsubramani, Hu,
  Yasunaga, Phillips, Gao, et~al.]{koh2021wilds}
Pang~Wei Koh, Shiori Sagawa, Henrik Marklund, Sang~Michael Xie, Marvin Zhang,
  Akshay Balsubramani, Weihua Hu, Michihiro Yasunaga, Richard~Lanas Phillips,
  Irena Gao, et~al.
\newblock Wilds: A benchmark of in-the-wild distribution shifts.
\newblock In \emph{International Conference on Machine Learning}, pp.\
  5637--5664, 2021.

\bibitem[Kouw \& Loog(2019)Kouw and Loog]{kouw2019review}
Wouter~M Kouw and Marco Loog.
\newblock A review of domain adaptation without target labels.
\newblock \emph{IEEE Transactions on Pattern Analysis and Machine
  Intelligence}, 43\penalty0 (3):\penalty0 766--785, 2019.

\bibitem[Krishnamachari et~al.(2023)Krishnamachari, Ng, and
  Foo]{krishnamachari2023mitigating}
Kiran Krishnamachari, See-Kiong Ng, and Chuan-Sheng Foo.
\newblock Mitigating real-world distribution shifts in the fourier domain.
\newblock \emph{Transactions on Machine Learning Research}, 2023.

\bibitem[Krizhevsky et~al.(2009)Krizhevsky, Hinton,
  et~al.]{krizhevsky2009learning}
Alex Krizhevsky, Geoffrey Hinton, et~al.
\newblock \emph{Learning multiple layers of features from tiny images}.
\newblock Citeseer, 2009.

\bibitem[Li et~al.(2022)Li, Li, Xiong, and Hoi]{li2022blip}
Junnan Li, Dongxu Li, Caiming Xiong, and Steven Hoi.
\newblock Blip: Bootstrapping language-image pre-training for unified
  vision-language understanding and generation.
\newblock In \emph{International Conference on Machine Learning}, 2022.

\bibitem[Li et~al.(2023)Li, Li, Savarese, and Hoi]{li2023blip2}
Junnan Li, Dongxu Li, Silvio Savarese, and Steven Hoi.
\newblock {BLIP-2:} bootstrapping language-image pre-training with frozen image
  encoders and large language models.
\newblock In \emph{International Conference on Machine Learning}, 2023.

\bibitem[Li et~al.(2021)Li, Lv, Xie, Liu, Liang, and Qin]{li2021bi}
Shuang Li, Fangrui Lv, Binhui Xie, Chi~Harold Liu, Jian Liang, and Chen Qin.
\newblock Bi-classifier determinacy maximization for unsupervised domain
  adaptation.
\newblock In \emph{Proceedings of the AAAI conference on artificial
  intelligence}, pp.\  8455--8464, 2021.

\bibitem[Liang et~al.(2018)Liang, Li, and Srikant]{liang2018enhancing}
Shiyu Liang, Yixuan Li, and Rayadurgam Srikant.
\newblock Enhancing the reliability of out-of-distribution image detection in
  neural networks.
\newblock In \emph{International Conference on Learning Representations}, 2018.

\bibitem[Liang et~al.(2019)Liang, Poggio, Rakhlin, and Stokes]{liang2019fisher}
Tengyuan Liang, Tomaso Poggio, Alexander Rakhlin, and James Stokes.
\newblock Fisher-rao metric, geometry, and complexity of neural networks.
\newblock In \emph{The 22nd International Conference on Artificial Intelligence
  and Statistics}, pp.\  888--896, 2019.

\bibitem[Lipton et~al.(2018)Lipton, Wang, and Smola]{lipton2018detecting}
Zachary Lipton, Yu-Xiang Wang, and Alexander Smola.
\newblock Detecting and correcting for label shift with black box predictors.
\newblock In \emph{International Conference on Machine Learning}, pp.\
  3122--3130, 2018.

\bibitem[Liu et~al.(2021)Liu, Haghgoo, Chen, Raghunathan, Koh, Sagawa, Liang,
  and Finn]{pmlr-v139-liu21f}
Evan~Z Liu, Behzad Haghgoo, Annie~S Chen, Aditi Raghunathan, Pang~Wei Koh,
  Shiori Sagawa, Percy Liang, and Chelsea Finn.
\newblock Just train twice: Improving group robustness without training group
  information.
\newblock In \emph{International Conference on Machine Learning}, pp.\
  6781--6792, 2021.

\bibitem[Lu et~al.(2024)Lu, Qin, Zhai, Shen, Chen, Wang, Kolouri, Stepputtis,
  Campbell, and Sycara]{lu2024characterizing}
Yuzhe Lu, Yilong Qin, Runtian Zhai, Andrew Shen, Ketong Chen, Zhenlin Wang,
  Soheil Kolouri, Simon Stepputtis, Joseph Campbell, and Katia Sycara.
\newblock Characterizing out-of-distribution error via optimal transport.
\newblock In \emph{Advances in Neural Information Processing Systems}, 2024.

\bibitem[Mansilla et~al.(2021)Mansilla, Echeveste, Milone, and
  Ferrante]{mansilla2021domain}
Lucas Mansilla, Rodrigo Echeveste, Diego~H Milone, and Enzo Ferrante.
\newblock Domain generalization via gradient surgery.
\newblock In \emph{Proceedings of the IEEE International Conference on Computer
  Vision}, 2021.

\bibitem[Mansour et~al.(2008)Mansour, Mohri, and
  Rostamizadeh]{mansour2008domain}
Yishay Mansour, Mehryar Mohri, and Afshin Rostamizadeh.
\newblock Domain adaptation with multiple sources.
\newblock In \emph{Advances in neural information processing systems}, 2008.

\bibitem[Mansour et~al.(2009)Mansour, Mohri, and
  Rostamizadeh]{mansour2009domain}
Yishay Mansour, Mehryar Mohri, and Afshin Rostamizadeh.
\newblock Domain adaptation: Learning bounds and algorithms.
\newblock \emph{arXiv preprint arXiv:0902.3430}, 2009.

\bibitem[Miller et~al.(2021)Miller, Taori, Raghunathan, Sagawa, Koh, Shankar,
  Liang, Carmon, and Schmidt]{miller2021accuracy}
John~P Miller, Rohan Taori, Aditi Raghunathan, Shiori Sagawa, Pang~Wei Koh,
  Vaishaal Shankar, Percy Liang, Yair Carmon, and Ludwig Schmidt.
\newblock Accuracy on the line: on the strong correlation between
  out-of-distribution and in-distribution generalization.
\newblock In \emph{International Conference on Machine Learning}, pp.\
  7721--7735, 2021.

\bibitem[Minderer et~al.(2021)Minderer, Djolonga, Romijnders, Hubis, Zhai,
  Houlsby, Tran, and Lucic]{minderer2021revisiting}
Matthias Minderer, Josip Djolonga, Rob Romijnders, Frances Hubis, Xiaohua Zhai,
  Neil Houlsby, Dustin Tran, and Mario Lucic.
\newblock Revisiting the calibration of modern neural networks.
\newblock In \emph{Advances in Neural Information Processing Systems}, pp.\
  15682--15694, 2021.

\bibitem[Nagarajan \& Kolter(2019)Nagarajan and
  Kolter]{nagarajan2019generalization}
Vaishnavh Nagarajan and J~Zico Kolter.
\newblock Generalization in deep networks: The role of distance from
  initialization.
\newblock \emph{arXiv preprint arXiv:1901.01672}, 2019.

\bibitem[Neyshabur et~al.(2015)Neyshabur, Tomioka, and
  Srebro]{neyshabur2015norm}
Behnam Neyshabur, Ryota Tomioka, and Nathan Srebro.
\newblock Norm-based capacity control in neural networks.
\newblock In \emph{Conference on Learning Theory}, pp.\  1376--1401, 2015.

\bibitem[Neyshabur et~al.(2017)Neyshabur, Bhojanapalli, McAllester, and
  Srebro]{neyshabur2017exploring}
Behnam Neyshabur, Srinadh Bhojanapalli, David McAllester, and Nati Srebro.
\newblock Exploring generalization in deep learning.
\newblock In \emph{Advances in Neural Information Processing systems}, 2017.

\bibitem[Neyshabur et~al.(2018)Neyshabur, Bhojanapalli, and
  Srebro]{neyshabur2017pac}
Behnam Neyshabur, Srinadh Bhojanapalli, and Nathan Srebro.
\newblock A pac-bayesian approach to spectrally-normalized margin bounds for
  neural networks.
\newblock In \emph{International Conference on Learning Representations}, 2018.

\bibitem[Ovadia et~al.(2019)Ovadia, Fertig, Ren, Nado, Sculley, Nowozin,
  Dillon, Lakshminarayanan, and Snoek]{ovadia2019can}
Yaniv Ovadia, Emily Fertig, Jie Ren, Zachary Nado, David Sculley, Sebastian
  Nowozin, Joshua Dillon, Balaji Lakshminarayanan, and Jasper Snoek.
\newblock Can you trust your model's uncertainty? evaluating predictive
  uncertainty under dataset shift.
\newblock In \emph{Advances in Neural Information Processing Systems}, 2019.

\bibitem[Peng et~al.(2019)Peng, Bai, Xia, Huang, Saenko, and
  Wang]{peng2019moment}
Xingchao Peng, Qinxun Bai, Xide Xia, Zijun Huang, Kate Saenko, and Bo~Wang.
\newblock Moment matching for multi-source domain adaptation.
\newblock In \emph{Proceedings of the IEEE International Conference on Computer
  Vision}, pp.\  1406--1415, 2019.

\bibitem[Qiao \& Peng(2021)Qiao and Peng]{qiao2021uncertainty}
Fengchun Qiao and Xi~Peng.
\newblock Uncertainty-guided model generalization to unseen domains.
\newblock In \emph{Proceedings of the IEEE Conference on Computer Vision and
  Pattern Recognition}, pp.\  6790--6800, 2021.

\bibitem[Radford et~al.(2021)Radford, Kim, Hallacy, Ramesh, Goh, Agarwal,
  Sastry, Askell, Mishkin, Clark, et~al.]{radford2021learning}
Alec Radford, Jong~Wook Kim, Chris Hallacy, Aditya Ramesh, Gabriel Goh,
  Sandhini Agarwal, Girish Sastry, Amanda Askell, Pamela Mishkin, Jack Clark,
  et~al.
\newblock Learning transferable visual models from natural language
  supervision.
\newblock In \emph{International Conference on Machine Learning}, 2021.

\bibitem[Recht et~al.(2018{\natexlab{a}})Recht, Roelofs, Schmidt, and
  Shankar]{recht2018cifar}
Benjamin Recht, Rebecca Roelofs, Ludwig Schmidt, and Vaishaal Shankar.
\newblock Do cifar-10 classifiers generalize to cifar-10?
\newblock \emph{arXiv preprint arXiv:1806.00451}, 2018{\natexlab{a}}.

\bibitem[Recht et~al.(2018{\natexlab{b}})Recht, Roelofs, Schmidt, and
  Shankar]{recht2018cifar10.1}
Benjamin Recht, Rebecca Roelofs, Ludwig Schmidt, and Vaishaal Shankar.
\newblock Do cifar-10 classifiers generalize to cifar-10?
\newblock \emph{arXiv preprint arXiv:1806.00451}, 2018{\natexlab{b}}.

\bibitem[Recht et~al.(2019)Recht, Roelofs, Schmidt, and
  Shankar]{recht2019imagenet}
Benjamin Recht, Rebecca Roelofs, Ludwig Schmidt, and Vaishaal Shankar.
\newblock Do imagenet classifiers generalize to imagenet?
\newblock In \emph{International Conference on Machine Learning}, pp.\
  5389--5400, 2019.

\bibitem[Sagawa et~al.(2020)Sagawa, Koh, Hashimoto, and
  Liang]{sagawa2019distributionally}
Shiori Sagawa, Pang~Wei Koh, Tatsunori~B Hashimoto, and Percy Liang.
\newblock Distributionally robust neural networks for group shifts: On the
  importance of regularization for worst-case generalization.
\newblock In \emph{International Conference on Learning Representations}, 2020.

\bibitem[Sagawa et~al.(2021)Sagawa, Koh, Lee, Gao, Xie, Shen, Kumar, Hu,
  Yasunaga, Marklund, et~al.]{sagawa2021extending}
Shiori Sagawa, Pang~Wei Koh, Tony Lee, Irena Gao, Sang~Michael Xie, Kendrick
  Shen, Ananya Kumar, Weihua Hu, Michihiro Yasunaga, Henrik Marklund, et~al.
\newblock Extending the wilds benchmark for unsupervised adaptation.
\newblock \emph{arXiv preprint arXiv:2112.05090}, 2021.

\bibitem[Schiff et~al.(2021)Schiff, Quanz, Das, and Chen]{schiff2021predicting}
Yair Schiff, Brian Quanz, Payel Das, and Pin-Yu Chen.
\newblock Predicting deep neural network generalization with perturbation
  response curves.
\newblock In \emph{Advances in Neural Information Processing Systems}, pp.\
  21176--21188, 2021.

\bibitem[Shi et~al.(2021)Shi, Seely, Torr, Siddharth, Hannun, Usunier, and
  Synnaeve]{shi2021gradient}
Yuge Shi, Jeffrey Seely, Philip~HS Torr, N~Siddharth, Awni Hannun, Nicolas
  Usunier, and Gabriel Synnaeve.
\newblock Gradient matching for domain generalization.
\newblock \emph{arXiv preprint arXiv:2104.09937}, 2021.

\bibitem[Simonyan \& Zisserman(2014)Simonyan and Zisserman]{simonyan2014very}
Karen Simonyan and Andrew Zisserman.
\newblock Very deep convolutional networks for large-scale image recognition.
\newblock \emph{arXiv preprint arXiv:1409.1556}, 2014.

\bibitem[Singh et~al.(2022)Singh, Hu, Goswami, Couairon, Galuba, Rohrbach, and
  Kiela]{singh2022flava}
Amanpreet Singh, Ronghang Hu, Vedanuj Goswami, Guillaume Couairon, Wojciech
  Galuba, Marcus Rohrbach, and Douwe Kiela.
\newblock Flava: A foundational language and vision alignment model.
\newblock In \emph{Proceedings of the IEEE/CVF Conference on Computer Vision
  and Pattern Recognition}, 2022.

\bibitem[Smith \& Le(2018)Smith and Le]{smith2017bayesian}
Samuel~L Smith and Quoc~V Le.
\newblock A bayesian perspective on generalization and stochastic gradient
  descent.
\newblock In \emph{International Conference on Learning Representations}, 2018.

\bibitem[Sun \& Saenko(2016)Sun and Saenko]{sun2016deep}
Baochen Sun and Kate Saenko.
\newblock Deep coral: Correlation alignment for deep domain adaptation.
\newblock In \emph{European Conference on Computer Vision}, pp.\  443--450,
  2016.

\bibitem[Sun et~al.(2022)Sun, Lu, and Ling]{sun2022prior}
Tao Sun, Cheng Lu, and Haibin Ling.
\newblock Prior knowledge guided unsupervised domain adaptation.
\newblock In \emph{European Conference on Computer Vision}, 2022.

\bibitem[Tachet~des Combes et~al.(2020)Tachet~des Combes, Zhao, Wang, and
  Gordon]{tachet2020domain}
Remi Tachet~des Combes, Han Zhao, Yu-Xiang Wang, and Geoffrey~J Gordon.
\newblock Domain adaptation with conditional distribution matching and
  generalized label shift.
\newblock In \emph{Advances in Neural Information Processing Systems}, pp.\
  19276--19289, 2020.

\bibitem[Tan \& Le(2019)Tan and Le]{tan2019efficientnet}
Mingxing Tan and Quoc Le.
\newblock Efficientnet: Rethinking model scaling for convolutional neural
  networks.
\newblock In \emph{International Conference on Machine Learning}, pp.\
  6105--6114, 2019.

\bibitem[Teney et~al.(2024)Teney, Lin, Oh, and Abbasnejad]{teney2024id}
Damien Teney, Yong Lin, Seong~Joon Oh, and Ehsan Abbasnejad.
\newblock Id and ood performance are sometimes inversely correlated on
  real-world datasets.
\newblock In \emph{Advances in Neural Information Processing Systems}, 2024.

\bibitem[Tu et~al.(2023)Tu, Deng, Gedeon, and Zheng]{Tu_2023_CVPR}
Weijie Tu, Weijian Deng, Tom Gedeon, and Liang Zheng.
\newblock A bag-of-prototypes representation for dataset-level applications.
\newblock In \emph{Proceedings of the IEEE/CVF Conference on Computer Vision
  and Pattern Recognition}, 2023.

\bibitem[Vedantam et~al.(2021)Vedantam, Lopez-Paz, and
  Schwab]{vedantam2021empirical}
Ramakrishna Vedantam, David Lopez-Paz, and David~J Schwab.
\newblock An empirical investigation of domain generalization with empirical
  risk minimizers.
\newblock In \emph{Advances in Neural Information Processing Systems}, pp.\
  28131--28143, 2021.

\bibitem[Volpi et~al.(2018)Volpi, Namkoong, Sener, Duchi, Murino, and
  Savarese]{volpi2018generalizing}
Riccardo Volpi, Hongseok Namkoong, Ozan Sener, John Duchi, Vittorio Murino, and
  Silvio Savarese.
\newblock Generalizing to unseen domains via adversarial data augmentation.
\newblock In \emph{Advances in Neural Information Processing Systems}, 2018.

\bibitem[Wang et~al.(2019)Wang, Ge, Lipton, and Xing]{wang2019learning}
Haohan Wang, Songwei Ge, Zachary Lipton, and Eric~P Xing.
\newblock Learning robust global representations by penalizing local predictive
  power.
\newblock In \emph{Advances in Neural Information Processing Systems}, 2019.

\bibitem[Wang \& Isola(2020)Wang and Isola]{wang2020understanding}
Tongzhou Wang and Phillip Isola.
\newblock Understanding contrastive representation learning through alignment
  and uniformity on the hypersphere.
\newblock In \emph{International Conference on Machine Learning}, pp.\
  9929--9939, 2020.

\bibitem[Wenzel et~al.(2022)Wenzel, Dittadi, Gehler, Simon-Gabriel, Horn,
  Zietlow, Kernert, Russell, Brox, Schiele, et~al.]{wenzel2022assaying}
Florian Wenzel, Andrea Dittadi, Peter~Vincent Gehler, Carl-Johann
  Simon-Gabriel, Max Horn, Dominik Zietlow, David Kernert, Chris Russell,
  Thomas Brox, Bernt Schiele, et~al.
\newblock Assaying out-of-distribution generalization in transfer learning.
\newblock In \emph{Advances in Neural Information Processing Systems}, 2022.

\bibitem[Wightman(2017)]{kl2017cifar}
Ross Wightman.
\newblock Train cifar10 with pytorch.
\newblock \url{https: //github.com/kuangliu/pytorch-cifar}, 2017.

\bibitem[Wightman(2019)]{rw2019timm}
Ross Wightman.
\newblock Pytorch image models.
\newblock \url{https://github.com/rwightman/pytorch-image-models}, 2019.

\bibitem[Xie et~al.(2024)Xie, Wei, Feng, Cao, and An]{xie2024importance}
Renchunzi Xie, Hongxin Wei, Lei Feng, Yuzhou Cao, and Bo~An.
\newblock On the importance of feature separability in predicting
  out-of-distribution error.
\newblock In \emph{Advances in Neural Information Processing Systems}, 2024.

\bibitem[Yang et~al.(2021)Yang, van~de Weijer, Herranz, Jui,
  et~al.]{yang2021exploiting}
Shiqi Yang, Joost van~de Weijer, Luis Herranz, Shangling Jui, et~al.
\newblock Exploiting the intrinsic neighborhood structure for source-free
  domain adaptation.
\newblock In \emph{Advances in Neural Information Processing Systems}, pp.\
  29393--29405, 2021.

\bibitem[You et~al.(2021)You, Liu, Wang, and Long]{you2021logme}
Kaichao You, Yong Liu, Jianmin Wang, and Mingsheng Long.
\newblock Logme: Practical assessment of pre-trained models for transfer
  learning.
\newblock In \emph{International Conference on Machine Learning}, pp.\
  12133--12143, 2021.

\bibitem[Yu et~al.(2022)Yu, Yang, Wei, Ma, and Steinhardt]{yu2022predicting}
Yaodong Yu, Zitong Yang, Alexander Wei, Yi~Ma, and Jacob Steinhardt.
\newblock Predicting out-of-distribution error with the projection norm.
\newblock In \emph{International Conference on Machine Learning}, 2022.

\bibitem[Zhai et~al.(2023)Zhai, Mustafa, Kolesnikov, and
  Beyer]{zhai2023sigmoid}
Xiaohua Zhai, Basil Mustafa, Alexander Kolesnikov, and Lucas Beyer.
\newblock Sigmoid loss for language image pre-training.
\newblock In \emph{International Conference on Computer Vision}, 2023.

\bibitem[Zhang et~al.(2021)Zhang, Ahuja, Xu, Wang, and Courville]{zhang2021can}
Dinghuai Zhang, Kartik Ahuja, Yilun Xu, Yisen Wang, and Aaron Courville.
\newblock Can subnetwork structure be the key to out-of-distribution
  generalization?
\newblock In \emph{International Conference on Machine Learning}, pp.\
  12356--12367, 2021.

\bibitem[Zhang et~al.(2019)Zhang, Liu, Long, and Jordan]{zhang2019bridging}
Yuchen Zhang, Tianle Liu, Mingsheng Long, and Michael Jordan.
\newblock Bridging theory and algorithm for domain adaptation.
\newblock In \emph{International Conference on Machine Learning}, pp.\
  7404--7413, 2019.

\bibitem[Zhao et~al.(2020)Zhao, Liu, Peng, and Metaxas]{zhao2020maximum}
Long Zhao, Ting Liu, Xi~Peng, and Dimitris Metaxas.
\newblock Maximum-entropy adversarial data augmentation for improved
  generalization and robustness.
\newblock In \emph{Advances in Neural Information Processing Systems}, 2020.

\bibitem[Zhou et~al.(2022)Zhou, Liu, Qiao, Xiang, and Loy]{zhou2022domain}
Kaiyang Zhou, Ziwei Liu, Yu~Qiao, Tao Xiang, and Chen~Change Loy.
\newblock Domain generalization: A survey.
\newblock \emph{IEEE Transactions on Pattern Analysis and Machine
  Intelligence}, 2022.

\end{thebibliography}
\bibliographystyle{tmlr}

\newpage
\appendix
\section{Appendix}
In the appendix, we first introduce the experimental details including access to models, datasets, and computation resources.

\subsection{Experimental Setup}
\subsubsection{Datasets} 
We carefully check the licenses of all datasets used in the experiment and list the open sources to them.

\noindent \texttt{ImageNet} \citep{deng2009imagenet}: \\ (\textcolor{blue}{https://www.image-net.org}); \\
\texttt{ImageNet-C} \citep{hendrycks2019benchmarking}: \\ (\textcolor{blue}{https://github.com/hendrycks/robustness});\\
\texttt{ImageNet-V2} \citep{recht2019imagenet}: \\ (\textcolor{blue}{https://github.com/modestyachts/ImageNetV2});\\
\texttt{ImageNet-Adversarial} \citep{hendrycks2021natural}: \\  (\textcolor{blue}{https://github.com/hendrycks/natural-adv-examples});\\
\texttt{ImageNet-Rendition} \citep{hendrycks2021many}: \\  (\textcolor{blue}{https://github.com/hendrycks/robustness}); \\
\texttt{ImageNet-Sketch} \citep{wang2019learning}: \\  (\textcolor{blue}{https://github.com/HaohanWang/ImageNet-Sketch});\\
\texttt{ObjectNet} \citep{barbu2019objectnet}: \\  (\textcolor{blue}{https://objectnet.dev/download.html});\\
\texttt{CIFAR-10} \citep{krizhevsky2009learning}: \\  (\textcolor{blue}{https://www.cs.toronto.edu/~kriz/cifar.html});\\
\texttt{CIFAR-10-C} \citep{hendrycks2019benchmarking}: \\  (\textcolor{blue}{https://github.com/hendrycks/robustness}); \\
\texttt{CIFAR-10.2} \citep{recht2018cifar}: \\  (\textcolor{blue}{https://github.com/modestyachts/CIFAR-10.1});\\
\texttt{CINIC} \citep{darlow2018cinic}: \\ (\textcolor{blue}{https://www.v7labs.com/open-datasets/cinic-10});\\
\texttt{Camelyon17} \citep{bandi2018detection}: \\ (\textcolor{blue}{https://camelyon17.grand-challenge.org/});\\
\texttt{DomainNet} \citep{peng2019moment}: \\  (\textcolor{blue}{http://ai.bu.edu/M3SDA/});\\

\subsubsection{Model Pool}
\noindent \textbf{ImageNet models} are publicly available via TIMM \citep{rw2019timm}. Models pre-trained by contrastive learning can be used from model-vs-human project \citep{geirhos2021partial}. Some MAE models \citep{he2022masked} are accessible from (\textcolor{blue}{https://github.com/facebookresearch/mae}).\\
\texttt{PIRL} \\ 
\texttt{InsDis} \\ 
\texttt{MoCo} \\ 
\texttt{MoCoV2} \\ 
\texttt{InfoMin} \\ 
\texttt{simclr\_resnet50x4} \\ 
\texttt{simclr\_resnet50x1} \\ 
\texttt{simclr\_resnet50x2} \\ 
\texttt{resnet50\_l2\_eps1} \\ 
\texttt{resnet50\_l2\_eps0\_5} \\ 
\texttt{resnet50\_l2\_eps0\_25} \\ 
\texttt{resnet50\_l2\_eps0\_03} \\ 
\texttt{resnet50\_l2\_eps0\_01} \\ 
\texttt{ens\_adv\_inception\_resnet\_v2} \\ 
\texttt{adv\_inception\_v3} \\ 
\texttt{tf\_efficientnet\_b0\_ap} \\ 
\texttt{resnet50\_trained\_on\_SIN\_and\_IN} \\ 
\texttt{resnet50\_trained\_on\_SIN\_}
\texttt{and\_IN\_then\_finetuned\_on\_IN} \\ 
\texttt{resnet50\_trained\_on\_SIN} \\ 
\texttt{resnet50\_augmix} \\ 
\texttt{resnet50\_cutmix} \\ 
\texttt{resnet50\_deepaugment} \\ 
\texttt{resnet50\_deepaugment\_and\_augmix} \\ 
\texttt{resnet50\_feature\_cutmix} \\ 
\texttt{resnet50\_pixmix} \\ 
\texttt{convnext\_xlarge\_384\_in22ft1k} \\ 
\texttt{convnext\_xlarge\_in22ft1k} \\ 
\texttt{resnetv2\_152x2\_bitm} \\ 
\texttt{resnetv2\_50x1\_bitm} \\ 
\texttt{convnext\_base\_384\_in22ft1k} \\ 
\texttt{convnext\_base\_in22ft1k} \\ 
\texttt{resmlp\_big\_24\_224\_in22ft1k} \\ 
\texttt{resmlp\_big\_24\_distilled\_224} \\ 
\texttt{tf\_efficientnetv2\_s\_in21ft1k} \\ 
\texttt{tf\_efficientnetv2\_m\_in21ft1k} \\ 
\texttt{tf\_efficientnetv2\_l\_in21ft1k} \\ 
\texttt{tf\_efficientnetv2\_xl\_in21ft1k} \\ 
\texttt{vit\_large\_patch16\_384} \\ 
\texttt{swin\_large\_patch4\_window12\_384} \\ 
\texttt{beit\_large\_patch16\_512} \\ 
\texttt{beit\_large\_patch16\_384} \\ 
\texttt{beit\_large\_patch16\_224} \\ 
\texttt{beit\_base\_patch16\_384} \\ 
\texttt{vit\_base\_patch16\_384} \\ 
\texttt{vit\_small\_r26\_s32\_384} \\ 
\texttt{vit\_tiny\_patch16\_384} \\ 
\texttt{vit\_large\_r50\_s32\_384} \\ 
\texttt{mixer\_b16\_224\_miil} \\ 
\texttt{resmlp\_big\_24\_224} \\ 
\texttt{resnetv2\_50x1\_bit\_distilled} \\ 
\texttt{ig\_resnext101\_32x16d} \\ 
\texttt{ig\_resnext101\_32x32d} \\ 
\texttt{ig\_resnext101\_32x8d} \\ 
\texttt{ig\_resnext101\_32x48d} \\ 
\texttt{resnext101\_32x16d\_wsl} \\ 
\texttt{tf\_efficientnet\_l2\_ns\_475} \\ 
\texttt{tf\_efficientnet\_b7\_ns} \\ 
\texttt{tf\_efficientnet\_b6\_ns} \\ 
\texttt{tf\_efficientnet\_b5\_ns} \\ 
\texttt{ssl\_resnext101\_32x8d} \\ 
\texttt{ssl\_resnext101\_32x16d} \\ 
\texttt{swsl\_resnext101\_32x8d} \\ 
\texttt{swsl\_resnext101\_32x16d} \\ 
\texttt{ssl\_resnext101\_32x4d} \\ 
\texttt{ssl\_resnext50\_32x4d} \\ 
\texttt{ssl\_resnet50} \\ 
\texttt{swsl\_resnext101\_32x4d} \\ 
\texttt{swsl\_resnext50\_32x4d} \\ 
\texttt{swsl\_resnet50} \\ 
\texttt{swin\_small\_patch4\_window7\_224} \\ 
\texttt{swin\_base\_patch4\_window12\_384} \\ 
\texttt{deit\_base\_patch16\_224} \\ 
\texttt{deit\_small\_distilled\_patch16\_224} \\ 
\texttt{resmlp\_36\_224} \\ 
\texttt{cait\_s36\_384} \\ 
\texttt{cait\_s24\_224} \\ 
\texttt{convit\_base} \\ 
\texttt{convit\_tiny} \\ 
\texttt{twins\_pcpvt\_base} \\ 
\texttt{eca\_nfnet\_l1} \\ 
\texttt{xcit\_tiny\_24\_p8\_384\_dist} \\ 
\texttt{efficientnet\_b1} \\ 
\texttt{efficientnet\_b3} \\ 
\texttt{efficientnet\_b4} \\ 
\texttt{tf\_efficientnet\_b2} \\ 
\texttt{tf\_efficientnet\_lite1} \\ 
\texttt{convnext\_base} \\ 
\texttt{convnext\_small} \\ 
\texttt{resnetrs350} \\ 
\texttt{pit\_xs\_distilled\_224} \\ 
\texttt{crossvit\_small\_240} \\ 
\texttt{botnet26t\_256} \\ 
\texttt{tinynet\_e} \\ 
\texttt{tinynet\_d} \\ 
\texttt{repvgg\_b2g4} \\ 
\texttt{mnasnet\_small} \\ 
\texttt{dla46x\_c} \\ 
\texttt{lcnet\_050} \\ 
\texttt{tv\_resnet34} \\ 
\texttt{tv\_resnet50} \\ 
\texttt{tv\_resnet101} \\ 
\texttt{tv\_resnet152} \\ 
\texttt{densenet121} \\ 
\texttt{inception\_v4} \\ 
\texttt{resnet26d} \\ 
\texttt{mobilenetv2\_140} \\ 
\texttt{hrnet\_w40} \\ 
\texttt{xception} \\ 
\texttt{xception41} \\ 
\texttt{resnet18} \\ 
\texttt{resnet34} \\ 
\texttt{seresnet50} \\ 
\texttt{mobilenetv2\_050} \\ 
\texttt{seresnet33ts} \\ 
\texttt{wide\_resnet50\_2} \\ 
\texttt{wide\_resnet101\_2} \\ 
\texttt{resnet18d} \\ 
\texttt{hrnet\_w18\_small} \\ 
\texttt{gluon\_resnet152\_v1d} \\ 
\texttt{hrnet\_w48} \\ 
\texttt{hrnet\_w44} \\ 
\texttt{repvgg\_b2} \\ 
\texttt{densenet201} \\ 
\texttt{hrnet\_w18\_small} \\ 
\texttt{resnet101d} \\ 
\texttt{gluon\_resnet101\_v1d} \\ 
\texttt{gluon\_resnet101\_v1s} \\ 
\texttt{gluon\_xception65} \\ 
\texttt{gluon\_seresnext50\_32x4d} \\ 
\texttt{gluon\_senet154} \\ 
\texttt{gluon\_inception\_v3} \\ 
\texttt{gluon\_resnet101\_v1c} \\ 
\texttt{tf\_inception\_v3} \\ 
\texttt{tv\_densenet121} \\ 
\texttt{tv\_resnext50\_32x4d} \\ 
\texttt{repvgg\_b1g4} \\ 
\texttt{resnext26ts} \\ 
\texttt{ghostnet\_100} \\ 
\texttt{crossvit\_9\_240} \\ 
\texttt{deit\_base\_patch16\_384} \\ 
\texttt{rexnet\_150} \\ 
\texttt{rexnet\_130} \\ 
\texttt{resnetrs50} \\ 
\texttt{resnet50d} \\ 
\texttt{resnet50} \\ 
\texttt{resnetv2\_50} \\ 
\texttt{resnetrs152} \\ 
\texttt{resnetrs101} \\ 
\texttt{resnet50\_aa} \\ 
\texttt{resnet50\_fastaa} \\ 
\texttt{resnet50\_randaa} \\ 
\texttt{vgg19\_bn} \\ 
\texttt{vgg16\_bn} \\ 
\texttt{vgg13\_bn} \\ 
\texttt{vgg11\_bn} \\ 
\texttt{vgg11} \\ 
\texttt{vgg11\_bn} \\ 
\texttt{vgg16} \\ 
\texttt{vgg19} \\ 
\texttt{resnet10t} \\ 
\texttt{resnet14t} \\ 
\texttt{darknet53} \\ 
\texttt{cs3darknet\_m} \\ 
\texttt{cs3darknet\_focus\_m} \\ 
\texttt{cs3darknet\_l} \\ 
\texttt{cs3darknet\_focus\_l} \\ 
\texttt{regnety\_040} \\ 
\texttt{regnety\_064} \\ 
\texttt{regnety\_080} \\ 
\texttt{regnetv\_040} \\ 
\texttt{regnetv\_064} \\ 
\texttt{regnetz\_040} \\ 
\texttt{regnetz\_040h} \\

\noindent \textbf{CIFAR-10 models} can be downloaded through (\textcolor{blue}{https://github.com/chenyaofo/pytorch-cifar-models}). The training script for training trajectory experiment borrows from (\textcolor{blue}{https://github.com/kuangliu/pytorch-cifar}).\\ 
\texttt{VGG16} \\ 
\texttt{VGG13} \\ 
\texttt{ResNet34} \\ 
\texttt{ResNet50} \\ 
\texttt{ResNet101} \\ 
\texttt{ResNet152} \\ 
\texttt{ShuffleNetG2} \\ 
\texttt{ShuffleNetG3} \\ 
\texttt{PreActResNet18} \\ 
\texttt{PreActResNet34} \\ 
\texttt{PreActResNet50} \\ 
\texttt{PreActResNet101} \\ 
\texttt{densenet\_cifar} \\ 
\texttt{DenseNet121} \\ 
\texttt{DenseNet169} \\ 
\texttt{DenseNet201} \\ 
\texttt{DenseNet161} \\ 
\texttt{ResNeXt29\_8x64d} \\ 
\texttt{ResNeXt29\_32x4d} \\ 
\texttt{MobileNetV2} \\ 
\texttt{RegNetX\_200MF} \\ 
\texttt{DLA} \\ 
\texttt{DPN} \\ 
\texttt{PNASNetB} \\ 
\texttt{RegNetX\_400MF} \\ 
\texttt{RegNetY\_400MF} \\ 
\texttt{MobileNet} \\ 
\texttt{ResNet18} \\ 
\texttt{VGG11} \\ 
\texttt{SimpleDLA} \\ 
\texttt{ResNeXt29\_4x64d} \\ 
\texttt{ResNeXt29\_2x64d} \\ 
\texttt{EfficientNetB0} \\ 
\texttt{SENet18} \\ 
\texttt{ShuffleNetV2} \\ 
\texttt{GoogLeNet} \\ 
\texttt{DPN92} \\ 
\texttt{ResNet34-160} \\ 
\texttt{ResNet34-170} \\ 
\texttt{ShuffleNetG2-170} \\ 
\texttt{ShuffleNetG2-180} \\ 
\texttt{ShuffleNetG2-190} \\ 
\texttt{SimpleDLA-90} \\ 
\texttt{SimpleDLA-105} \\ 
\texttt{SimpleDLA-120} \\ 
\texttt{SimpleDLA-135} \\ 
\texttt{SimpleDLA-150} \\ 
\texttt{SENet18-60} \\ 
\texttt{SENet18-75} \\ 
\texttt{SENet18-90} \\ 
\texttt{SENet18-120} \\ 
\texttt{SENet18-135} \\ 
\texttt{SENet18-150} \\ 
\texttt{SENet18-165} \\ 
\texttt{ShuffleNetG2-170} \\ 
\texttt{ShuffleNetG2-180} \\ 
\texttt{ShuffleNetG2-190} \\ 
\texttt{densenet\_cifar-135} \\ 
\texttt{densenet\_cifar-150} \\ 
\texttt{densenet\_cifar-165} \\ 
\texttt{MobileNetV2-135} \\ 
\texttt{MobileNetV2-150} \\ 
\texttt{MobileNetV2-165} \\ 
\texttt{LeNet} \\ 
\texttt{PreActResNet152} \\

\noindent \textbf{WILDS models} (Camelyon17 and DomainNet) are trained by github (\textcolor{blue}{https://github.com/p-lambda/wilds}).\\ 

\noindent \textbf{Zero-shot vision-language models} are provided in OpenCLIP \citep{ilharco_gabriel_2021_5143773}. They are listed as follows in the pattern (\texttt{architecture}, \texttt{source}):.\\ 
(\texttt{RN50},~\texttt{openai}) \\ 
(\texttt{RN50},~\texttt{yfcc15m}) \\ 
(\texttt{RN50},~\texttt{cc12m}) \\ 
(\texttt{RN50-quickgelu},~\texttt{yfcc15m}) \\ 
(\texttt{RN50-quickgelu},~\texttt{cc12m}) \\ 
(\texttt{RN101},~\texttt{openai}) \\ 
(\texttt{RN101},~\texttt{yfcc15m}) \\ 
(\texttt{RN101-quickgelu},~\texttt{yfcc15m}) \\ 
(\texttt{RN50$\times$4},~\texttt{openai}) \\ 
(\texttt{RN50$\times$16},~\texttt{openai}) \\ 
(\texttt{RN50$\times$64},~\texttt{openai}) \\ 
(\texttt{ViT-B-32},~\texttt{openai}) \\ 
(\texttt{ViT-B-32},~\texttt{laion400m\_e32}) \\ 
(\texttt{ViT-B-16},~\texttt{openai}) \\ 
(\texttt{ViT-B-16},~\texttt{laion400m\_e32}) \\ 
(\texttt{ViT-L-14},~\texttt{openai}) \\ 
(\texttt{ViT-L-14},~\texttt{laion2b\_s32b\_b82k}) \\ 
(\texttt{ViT-L-14-336},~\texttt{openai}) \\ 
(\texttt{ViT-H-14},~\texttt{laion2b\_s32b\_b79k}) \\ 
(\texttt{ViT-g-14},~\texttt{laion2b\_s34b\_b88k}) \\ 
(\texttt{ViT-bigG-14},~\texttt{laion2b\_s39b\_b160k}) \\ 
(\texttt{convnext\_base},~\texttt{laion400m\_s13b\_b51k}) \\ 
(\texttt{convnext\_base\_w},~\texttt{laion\_aesthetic\_s13b\_b82k}) \\ 
(\texttt{convnext\_base\_w\_320}, \texttt{laion\_aesthetic\_s13b\_b82k\_augreg}) \\ 
(\texttt{convnext\_large\_d},~\texttt{laion2b\_s26b\_b102k\_augreg}) \\ 
(\texttt{convnext\_large\_d\_320}, \texttt{laion2b\_s29b\_b131k\_ft\_soup}) \\ 
(\texttt{convnext\_xxlarge},~\texttt{laion2b\_s34b\_b82k\_augreg}) \\ 
(\texttt{ViT-B-32},
\texttt{Model-B-32\_Data-80M\_Samples-34B\_} 
\texttt{lr-1e-3\_bs-88k.pt}) \\ 
(\texttt{ViT-B-16},
\texttt{Model-B-16\_Data-80M\_Samples-34B\_} 
\texttt{lr-1e-3\_bs-88k.pt}) \\
(\texttt{ViT-L-14},
\texttt{Model-L-14\_Data-80M\_Samples-34B\_}
\texttt{lr-1e-3\_bs-88k.pt}) \\ 
(\texttt{ViT-B-32},~\texttt{datacomp\_m\_s128m\_b4k}) \\ 
(\texttt{ViT-B-32},~\texttt{commonpool\_m\_clip\_s128m\_b4k}) \\ 
(\texttt{ViT-B-32},~\texttt{commonpool\_m\_laion\_s128m\_b4k}) \\ 
(\texttt{ViT-B-32},~\texttt{commonpool\_m\_image\_s128m\_b4k}) \\ 
(\texttt{ViT-B-32},~\texttt{commonpool\_m\_text\_s128m\_b4k}) \\ 
(\texttt{ViT-B-32},~\texttt{commonpool\_m\_basic\_s128m\_b4k}) \\ 
(\texttt{ViT-B-32},~\texttt{commonpool\_m\_s128m\_b4k}) \\ 
(\texttt{ViT-B-32},~\texttt{datacomp\_s\_s13m\_b4k}) \\ 
(\texttt{ViT-B-32},~\texttt{commonpool\_s\_clip\_s13m\_b4k}) \\ 
(\texttt{ViT-B-32},~\texttt{commonpool\_s\_laion\_s13m\_b4k}) \\ 
(\texttt{ViT-B-32},~\texttt{commonpool\_s\_image\_s13m\_b4k}) \\ 
(\texttt{ViT-B-32},~\texttt{commonpool\_s\_text\_s13m\_b4k}) \\ 
(\texttt{ViT-B-32},~\texttt{commonpool\_s\_basic\_s13m\_b4k}) \\ 
(\texttt{ViT-B-32},~\texttt{commonpool\_s\_s13m\_b4k}) \\ 
(\texttt{ViT-B-16},~\texttt{datacomp\_l\_s1b\_b8k}) \\ 
(\texttt{ViT-L-14},~\texttt{datacomp\_xl\_s13b\_b90k}) \\ 
(\texttt{EVA01-g-14},~\texttt{laion400m\_s11b\_b41k}) \\ 
(\texttt{EVA01-g-14-plus},~\texttt{merged2b\_s11b\_b114k}) \\ 
(\texttt{EVA02-B-16},~\texttt{merged2b\_s8b\_b131k}) \\ 
(\texttt{EVA02-L-14},~\texttt{merged2b\_s4b\_b131k}) \\ 
(\texttt{EVA02-L-14-336},~\texttt{merged2b\_s6b\_b61k}) \\ 
(\texttt{EVA02-E-14},~\texttt{laion2b\_s4b\_b115k}) \\ 
(\texttt{EVA02-E-14-plus},~\texttt{laion2b\_s9b\_b144k}) \\ 
(\texttt{ViT-B-32-quickgelu},~\texttt{metaclip\_fullcc}) \\ 
(\texttt{ViT-B-16-quickgelu},~\texttt{metaclip\_fullcc}) \\ 
(\texttt{ViT-L-14-quickgelu},~\texttt{metaclip\_fullcc}) \\ 
(\texttt{ViT-L-14-CLIPA-336},~\texttt{datacomp1b}) \\ 
(\texttt{ViT-H-14-CLIPA},~\texttt{datacomp1b}) \\ 
(\texttt{ViT-H-14-CLIPA-336},~\texttt{datacomp1b}) \\ 
(\texttt{ViT-bigG-14-CLIPA},~\texttt{datacomp1b}) \\ 
(\texttt{ViT-bigG-14-CLIPA-336},~\texttt{datacomp1b}) \\ 
(\texttt{ViT-B-16-SigLIP},~\texttt{webli}) \\ 
(\texttt{ViT-B-16-SigLIP-256},~\texttt{webli}) \\ 
(\texttt{ViT-L-16-SigLIP-384},~\texttt{webli}) \\ 
(\texttt{ViT-SO400M-14-SigLIP},~\texttt{webli}) \\ 
(\texttt{ViT-SO400M-14-SigLIP-384},~\texttt{webli}) \\ 
(\texttt{ViT-B-32-quickgelu},~\texttt{metaclip\_fullcc}) \\ 
(\texttt{ViT-B-16-quickgelu},~\texttt{metaclip\_fullcc}) \\ 
(\texttt{ViT-L-14-quickgelu},~\texttt{metaclip\_fullcc}) \\ 

\noindent BLIP \citep{li2022blip} and BLIP-2 \citep{li2023blip2} models are from LAVIS library (\textcolor{blue}{https://github.com/salesforce/LAVIS/tree/main});\\

\subsubsection{Computation Resources} 
PyTorch version is \textcolor{black}{1.10.2+cu102}.
All experiments run on one 2080Ti and the CPU \textcolor{black}{AMD Ryzen Threadripper 2950X 16-Core Processor}.

\newpage

\revise{\subsection{Sensitivity analysis on $\tau_w$}
We use Scipy implementation of weighted Kendall's rank correlation ($\tau_w$)\footnote{https://docs.scipy.org/doc/scipy/reference/generated/scipy.stats.weightedtau.html}, which does not provide a p-value to indicate the statistical significance. It is because the null distribution of the statistic is unknown even in the additive hyperbolic case. Thus, to examine whether the evaluation metric is stable, we conduct a sensitivity analysis on $\tau_w$ by randomly dropping some models from the model pool. The results are summarized in Table \ref{tab:tau_soft} and Table \ref{tab:tau_conf}. We find that $\tau_w$ has a consistent assessment on two methods with different numbers of models, suggesting the stability of $\tau_w$.

}

\begin{table}
\begin{minipage}{.5\linewidth}
\small
    \centering
            \setlength{\tabcolsep}{2pt}
        \begin{tabular}{c c c c c c c}
        \toprule
         \textbf{Models} &  \multicolumn{1}{c}{\textbf{{ImageNet-V2-A}}} & \multicolumn{1}{c}{\textbf{{ImageNet-R}}}& \multicolumn{1}{c}{\textbf{{ObjectNet}}}\\ 
        \midrule
         $20\%$ & $0.705$ & $0.940$ & $0.856$\\
         $50\%$ & $0.746$ & $0.934$ & $0.894$\\
         $70\%$ & $0.778$ & $0.928$ & $0.906$\\
         $100\%$ & $0.758$ & $0.928$ & $0.895$\\
         \bottomrule
 	\end{tabular}
    \caption{Correlation strength of SoftmaxCorr with different number models.
    Weighted Kendall correlation ($\tau_w$) is used as the metric. 
    }
    \label{tab:tau_soft}
\end{minipage}
\quad
\begin{minipage}{.5\linewidth}
\small
    \centering
      \setlength{\tabcolsep}{2pt}
            \begin{tabular}{c c c c c c c}
        \toprule
         \textbf{Models} &  \multicolumn{1}{c}{\textbf{{ImageNet-V2-A}}} & \multicolumn{1}{c}{\textbf{{ImageNet-R}}}& \multicolumn{1}{c}{\textbf{{ObjectNet}}}\\ 
        \midrule
         $20\%$ & $0.540$ & $0.841$ & $0.786$ \\
         $50\%$ & $0.542$ & $0.877$ & $0.846$\\
         $70\%$ & $0.579$ & $0.867$ & $0.846$\\
         $100\%$ & $0.597$ & $0.884$ & $0.849$\\
         \bottomrule
 	\end{tabular}
    \caption{Correlation strength of MaxPred with different number models.
    Weighted Kendall correlation ($\tau_w$) is used as the metric. }
    \label{tab:tau_conf}
\end{minipage}
\end{table}

\end{document}